\documentclass[final,3p,times]{elsarticle}

\usepackage{booktabs}
\usepackage{natbib}
\usepackage{multirow}
\usepackage{amsfonts}
\usepackage{amssymb}
\usepackage{amsmath}
\usepackage[usenames,dvipsnames]{pstricks}
\usepackage{lscape}
\usepackage{rotating}
\usepackage{hyperref}
\usepackage{enumitem}
\usepackage{multirow}
\usepackage{color}
\usepackage{lscape}
\usepackage{colortbl}
\usepackage{url}
\usepackage{algorithm}
\usepackage{algorithmic}
\usepackage{lineno}
\usepackage{subcaption}
\usepackage{xfrac}

\usepackage[
    type={CC},
    modifier={by-nc-nd},
    version={4.0},
]{doclicense}

\usepackage{setspace} 

\newcommand{\tc}[2]{\textcolor [RGB]{0,0,0}{#2}}

\newcommand{\mytitle}{Detecting Mammals in UAV Images: Best Practices to address a substantially Imbalanced Dataset with Deep Learning}

\journal{Remote Sensing of Environment. DOI: 10.1016/j.rse.2018.06.028}

\begin{document}
\doclicenseThis
\begin{frontmatter}

\title{\mytitle}
\author[1]{Benjamin Kellenberger}
\author[1]{Diego Marcos}
\author[1]{Devis Tuia}
\cortext[grant]{Corresponding Author: BK, benjamin.kellenberger@wur.nl.}

\address[1]{Laboratory of Geo-Information Science and Remote Sensing, Wageningen University \& Research, the Netherlands}

\begin{abstract}
Knowledge over the number of animals in large wildlife reserves is a vital necessity for park rangers in their efforts to protect endangered species. Manual animal censuses are dangerous and expensive, hence Unmanned Aerial Vehicles (UAVs) with consumer level digital cameras are becoming a popular alternative tool to estimate livestock. Several works have been proposed that semi-automatically process UAV images to detect animals, of which some employ Convolutional Neural Networks (CNNs), a recent family of deep learning algorithms that proved very effective in object detection in large datasets from computer vision. However, the majority of works related to wildlife focuses only on small datasets (typically subsets of UAV campaigns)\tc{ and hence is likely to fail}{, which might be detrimental} when presented with the sheer scale of real study areas for large mammal census\tc{, in which most methods tend to generate}{. Methods may yield} thousands of false alarms \tc{}{in such cases}. In this paper, we study how to scale CNNs to large wildlife census tasks and present a number of recommendations to train a CNN on a large UAV dataset. We further introduce novel evaluation protocols that are tailored to censuses and model suitability for subsequent human verification of detections. Using our recommendations, we are able to train a CNN reducing the number of false positives by an order of magnitude \tc{}{compared to previous state-of-the-art}. Setting the requirements at 90\% recall, our CNN allows to reduce the \tc{effort}{amount of data} required for manual verification by three times, thus making it possible for rangers to screen all the data acquired efficiently and to detect almost all animals in the reserve automatically.
\end{abstract}


\end{frontmatter}


\section{Introduction}

Livestock censuses play an important part in the ever-ongoing fight against the rapid decline of endangered large mammal species~\citep{linchant2015unmanned}. Knowing the exact number of individuals as well as their last known location sheds light on environmental requirements for different species~\citep{gadiye2016spatial}, the developments of species reintroductions~\citep{berger2014using}, and can be of great help in anti-poaching efforts~\citep{piel2015deterring}.

Identifying and counting animals \tc{}{in remote areas} has traditionally been carried out manually, using methods such as surveys from manned \tc{helicopters}{aircrafts}~\citep{bayliss1989distribution,norton1978counting}\tc{ or}{,} camera traps~\citep{silver2004use} \tc{}{and other manual methods~\citep{jachmann1991evaluation}}. For a long time, such campaigns were the only means of getting rough estimations of animal abundances, but they come with substantial flaws: \emph{(i.)} they pose great risk on human operators who have to get close to armed poachers and wild animals; \emph{(ii.)} they are expensive, requiring many man-hours of surveying, and \emph{(iii.)} they \tc{are very limited in terms of the extent that can be monitored}{might lead to accuracy deficiencies due to the limited extents that can be monitored}. Camera trap\tc{-based surveys are limited to a small set of hot-spot areas~\citep{o2010camera},}{s have come down in cost~\citep{nazir2017wiseeye} but still require risky in-field installation and maintenance.} \tc{and manned helicopters may only scan small transects based on probability estimations of where animals could be~\citep{choquenot2008evaluating}}{Manned aircrafts overcome this problem, but are expensive and depend on human operators who might disagree and introduce estimation errors~\citep{schlossberg2016testing,bouche2012count}}. Therefore, \tc{even if these methods are more accurate, they still provide only rough approximations: the areas to be monitored, often as big as entire national parks and wildlife reserves, greatly exceed the feasible limits in size of manned surveys}{traditional methods generally lead to high monetary costs or raise safety concerns}. \tc{}{These factors are particularly limiting in remote areas like the African savanna examined in this study.}

A promising direction to address these issues is to employ Unmanned Aerial Vehicles (UAVs) for monitoring purposes \tc{}{\citep{hodgson2018}}. UAVs are remotely controlled, inexpensive aircrafts that can be equipped with sensing instruments such as point-and-shoot cameras. As such, UAVs alleviate both the risk on operators and the financial pressure on the data acquisition~\citep{linchant2015unmanned}. Furthermore, the bird's eye view allows to \tc{not only }{}reach otherwise inaccessible areas \tc{, but also to quickly scan extremely large areas in one go. While this enables the potential acquisition of entire national parks, it}{from a safe distance. However, bypassing human counting} inevitably \tc{shifts the problem of man hours to}{requires more time to be spent on} the analysis of the acquired data. Although the task of manual photo-interpretation does not expose the operator to the risks involved in field work, it can become prohibitively expensive for large UAV campaigns, which often generate tens of thousands of images. Works exist that employ experts to manually find animals in aerial images~\citep{diaz2017using,hodgson2013unmanned}, but they were indeed limited to small study areas.

This problem can be alleviated by automatically selecting and showing to the operator only the images that are most likely to contain an animal, which are typically a very small fraction of the total. \tc{O}{Although this is bound to introduce some false negatives, o}ne can leverage the improvements obtained in recent years in the computer vision field of object detection, which has already found a range of applications in remote sensing, \tc{}{to minimize their impact}. \tc{The target of}{Our objective is to use} machine-based object detection \tc{is}{} to train a model on a subset of the data that has been manually annotated and use it to predict the presence of objects in the rest of the dataset or in new acquisition campaigns. Object detectors work by extracting expressive image statistics (features) at locations of interest and using these features to classify each candidate location into a specific object class (i.e. ``animal'') or into a background class corresponding to the absence of any object of interest. Multiple approaches have been proposed to this end~\citep{ren2015faster,Redmon_2016_CVPR,NIPS2016_6465}, of which some have been applied to animal censuses. For example, several detectors can be found in the works of~\citet{Chamoso2014} and~\citet{van2014nature}, both tackling the problem of cattle counting. 
Similarly, \citet{Andrew_2017_ICCV} deploy deep object detectors and trackers to detect cattle, but using very low-flying UAVs.\newline 
Recent works consider the detection of large mammals in the African Savanna: \citet{yang2014spotting} and \citet{xue2017automatic} employ artificial neural networks on high resolution satellite imagery. \citet{Ofli2016} consider UAV data (the same data used in this study) and also provide a comprehensive set of annotations acquired by volunteers in a crowdsourcing setting. In~\citet{rey2017detecting}, authors build on that dataset and propose a classification system based on an ensemble of classifiers, each one specializing in detecting a single animal. We will use their methodology as a baseline for our experimentations.

Despite the theoretical advantage of detectors to process large datasets quickly, many of \tc{these}{such} works have been showcased on relatively small and confined study sites \tc{}{\citep{hollings2018}}, as for example \tc{the}{a} set of images containing at least one animal, or a balanced dataset with equal occurrences of the animals and background classes. Even if this is acceptable for academic exercises, it also has two consequences: \emph{(i.)} the small scale, and therefore the probable dataset autocorrelation, does not allow to assess the performance of the method, were it to be upscaled to much larger areas; \emph{(ii.)} the high concentration of individuals in the selected region hides the effect that covering vast, empty swaths of habitat has on the number of false positives that have then to be manually corrected. The few works that account for a realistic positive (animal) / negative (background) balance in a large mammal reserve~\citep{rey2017detecting,van2014nature} show that, at a recall rate of $\mathrm{80\%}$, at least 20 false positives should be expected for each true positive. This can have a big impact on the effort required for manual verification, limiting the advantage of using an automatic detector in the first place. Our aim is to reduce the manual effort required by leveraging state-of-the-art object detection models and exploring a number of proposed training recommendations to adapt such models to this challenging and extremely unbalanced setting.

As object detection models, we consider recent developments in deep learning. More specifically, we deploy convolutional neural networks (CNNs) for object detection~\citep{Redmon_2016_CVPR}. CNNs are nowadays the base building blocks of most computer vision pipelines and have also proven to be extremely successful in remote sensing~\citep{zhu2017deep}. Unlike traditional models, CNNs do not only train a classifier, but include the feature extraction part in the training process, which yields features that are of particular validity for the problem at hand. \tc{In practice, this means that CNNs can account for very subtle inter-class differences in appearance, while being invariant to intra-class differences. Such a behavior}{This} is of particular value in animal detection: animals show a substantial appearance heterogeneity due to their species, color and pose variations, and also due to external conditions such as motion blur, sensor differences, and different illumination settings. Furthermore, if seen from above, animals oftentimes tend to be hard to distinguish from various stationary objects like tree trunks, rocks, and dirt mounds (see Figure \ref{fig:appearanceConfusion} for examples). \tc{T}{Without end-to-end trained features, t}he result of this heterogeneity \tc{is that}{would increase the risk of} automated systems \tc{tend }{}to either miss animals because their appearance does not resemble what the model has learned (false negatives), or else falsely detect background objects that look like animals (false positives).

Summing up, we aim at taking a step towards solving the task of large mammal census in an African Savanna wildlife reserve using UAV still imagery and deep learning. \tc{The images have not been preselected for containing animals and represent the real proportion between animals and background. This tends to have the effect of canceling out the signals coming from the few true positives returned by automatic animal detectors with an order of magnitude more false positives. Since these have to be manually verified, the advantage of using an automatic detector is greatly impaired.}{} We \tc{show how using a CNN model with an appropriate training schedule can substantially reduce the number of false positives, almost by an order of magnitude, while keeping the same recall. This can remove the need to manually screen thousands of empty images and allows to scale the UAV census to whole reserves with minimal human supervision. }{present a series of recommendations that enable deep CNN models to address these issues. We showcase the recommendations on a dataset that features a realistic proportion between animals and background, which is challenging for detectors due to the small sample size and heavy class imbalance. Compared to the current baseline on our dataset~\citep{rey2017detecting}, a CNN trained with these recommendations is able to reduce the number of falsely detected animals by an order of magnitude, while still being able to score a high recall (up to 90\% of the animals present are detected). Moreover, since the (overall fewer) detections of our model are spread across a much lower number of UAV images, significantly less manual verification is required.} 


\section{Data}

\subsection{Study area and ground truth}

We base our studies on a dataset acquired over the Kuzikus wildlife reserve in eastern Namibia\footnote{\url{http://kuzikus-namibia.de/xe_index.html}}. Kuzikus is a private-owned park located at the edge of the Kalahari desert and covers an approximate area of $\mathrm{103km^2}$. According to the park's estimations the number of large mammals in the park exceeds 3000 individuals and consists of more than 20 species, such as Greater Kudus (\textit{Tragelaphus strepsiceros}), Gemsboks (\textit{Oryx gazella}), Hartebeests (\textit{Alcelaphus buselaphus}) and more (see  \citet{rey2017detecting}). Furthermore, Kuzikus is part of a national breeding project to enlarge the dwindling population of the Black Rhino (\textit{Diceros bicornis}).

The data were acquired between May 12 and May 15, 2014, by the SAVMAP Consortium\footnote{\url{http://lasig.epfl.ch/savmap}}. Five flight campaigns were conducted, and a lightweight single-wing UAV (SenseFly\footnote{\url{https://www.sensefly.com}} eBee), equipped with a Canon PowerShot S110 RGB camera, was employed. \tc{}{A multispectral and a thermal sensor were also used to acquire data, but their lower resolution, coupled with the low temperature contrast during the day, made the annotation of the animals on such data unfeasible.} During these campaigns the camera acquired a total of 654 images, each of size $\mathrm{3000\times4000}$ pixels and {24 bit} radiometric resolution. In total, this yielded more than 8.3 billion pixels, resp. an area of $\mathrm{13.38km^2}$, with an estimated average resolution of 4cm.

\begin{figure}
\centering
\begin{subfigure}[b]{.3\textwidth}
\includegraphics[width=\linewidth]{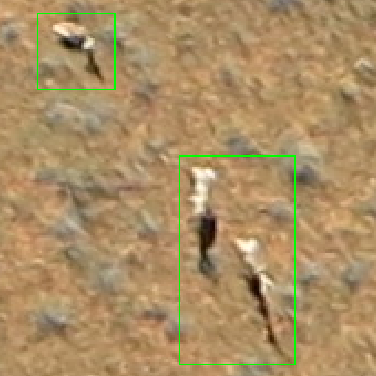}
\caption{}
\label{fig:anno_crowdsourced}
\end{subfigure}
\begin{subfigure}[b]{.3\textwidth}
\includegraphics[width=\linewidth]{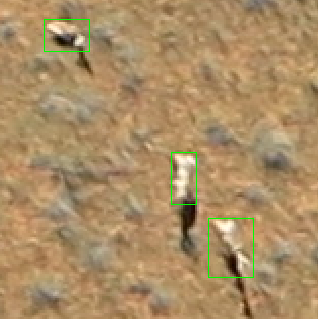}
\caption{}
\label{fig:anno_refined}
\end{subfigure}

\caption{An example of annotations retrieved by MicroMappers from the crowdsourcing campaign (left). Since the center points of each animal were used to train and test the models, all annotations had to be revised to only encompass individual animals themselves (right).}

\label{fig:annotation_refinement}

\end{figure}

\begin{figure}
\centering
\includegraphics[width=12cm]{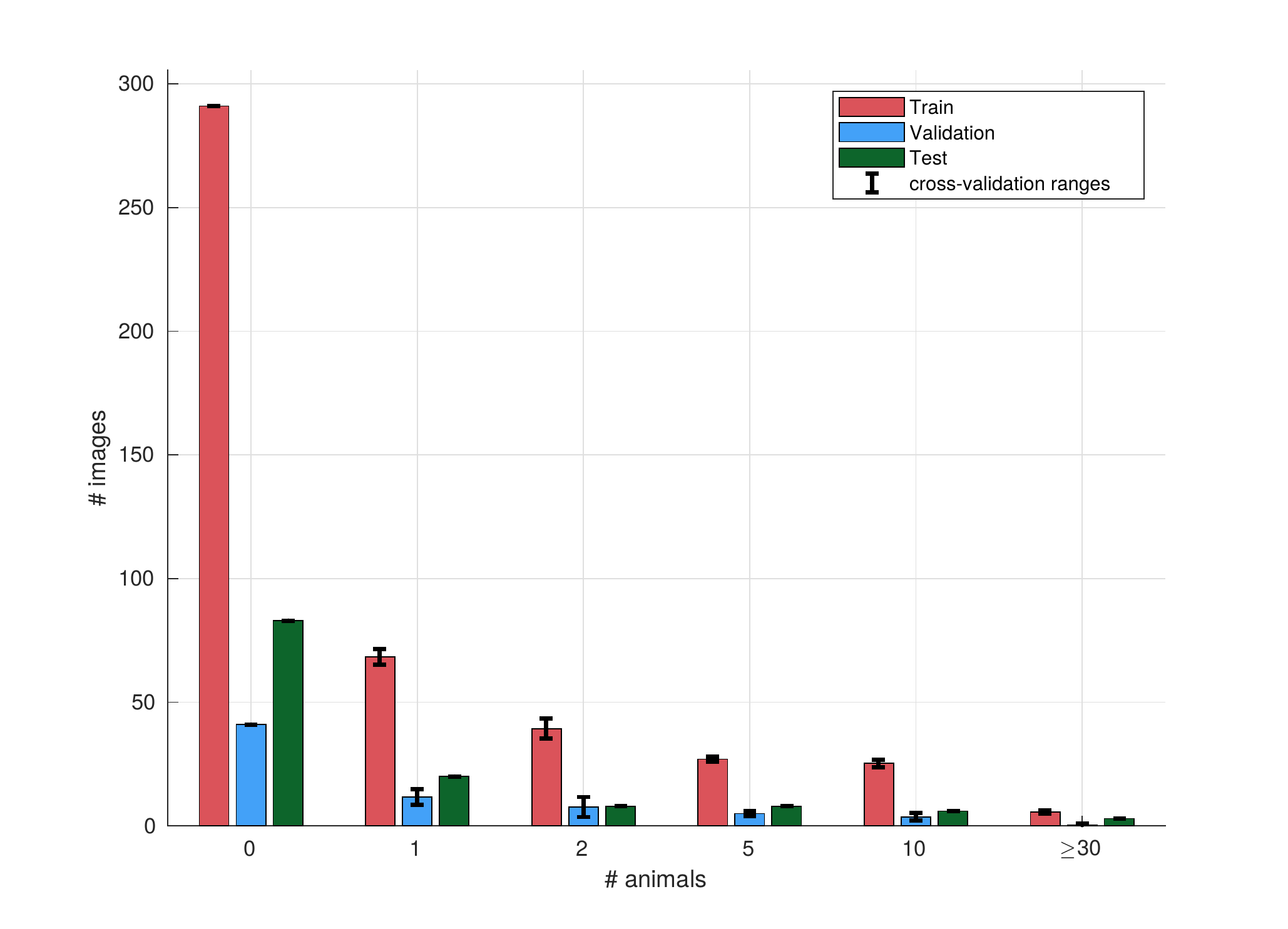}
\caption{Distribution of the number of animals per image for the training, validation and test sets. \tc{}{Standard deviation ranges are given for the three cross-validation splits.} The majority of images does not contain animals, which poses a significant challenge to any detector.}
\label{fig:datasetHistograms}
\end{figure}

An initial ground truth (convex hull polygons of large animals) was provided by MicroMappers\footnote{\url{https://micromappers.wordpress.com}}~\citep{Ofli2016} and consisted of 976 annotations in a subset of the 654 images\tc{}{, which was completed within three days}. Since they were based on crowdsourced volunteers efforts, some of the annotations were coarse or erroneous in that they occasionally omitted an animal, were not very accurate position-wise, or included multiple individuals in one annotation at once (Figure \ref{fig:anno_crowdsourced}). Even if the main target of this study was census-oriented, such blunders have detrimental effects on both the model's detection accuracy and the evaluation trustworthiness. For instance, sampling locations from the center point of a ground truth might result in the annotation lying between the animal and its shadow (i.e., on the ground). All annotations were thus screened and refined by the authors \tc{}{in another three days} to only include the animal itself, and only one animal per location (Figure \ref{fig:anno_refined}). \tc{}{We note at this point that due to the overlap between UAV images, it might occasionally be possible for one animal to appear in multiple images. Resolving this potential conflict would require matching heuristics that are beyond the scope of this paper.} For all training and evaluations, only the center locations of the annotations were retained. The data were then split into training (70\%), validation (10\%) and test (20\%) sets on an image-wise basis to ensure no annotation and no image were included in more than one set. Split priority was laid on the number of animals; all remaining empty images were likewise distributed according to the same percentages. \tc{}{Three different training and validation splits were created according to these rules, and the main models were trained on all three to avoid systematic biases due to the choice of splits. Due to the split rules and the uneven occurrence of animals in the images, the split statistics (number of images, number of animals, etc.) show slight fluctuations, but they are nevertheless within a reasonable range, of only a few percentage points, to each other. The test set is identical for the three cross-validation splits.} The final dataset statistics are summarized in Table \ref{tab:kuzikus_statistics} and Figure \ref{fig:datasetHistograms}. \tc{}{The data, including the annotations, will be made publicly available at \url{https://zenodo.org/record/16445}.}

\begin{table}
\centering
\begin{tabular}{r | r r r}
Set & \#pixels & \#images with/without animals & \#animals\\
\hline
training & $5.83*10^9$ ($7.3*10^7$) & 165.67 (6.5) / 291.00 (0) & 834.00 (5.3)\\
validation & $8.90*10^8$ ($7.3*10^7$) & 28.33 (6.5) / 41.00 (0) & 114.00 (5.3)\\
test & $1.65*10^9$ & 45/83 & 235\\
\hline
total & $8.36*10^9$ & 415/239 & 1183\\
\end{tabular}
\caption{Statistics of the Kuzikus 2014 acquisition. The data were split in an image-wise manner into roughly 70\% for training, 10\% for validation and 20\% for testing purposes, based on the number of animals. \tc{}{To test for potential data biases, three different training/validation splits were carried out; for the training and validation sets average numbers of images and animals, together with their standard deviation (in brackets), are provided. The test set was identical in all three runs.}}
\label{tab:kuzikus_statistics}
\end{table}

\subsection{The challenges of covering large areas}
\label{sec:largeAreasProblems}

\begin{figure}
\centering
\includegraphics[width=\textwidth]{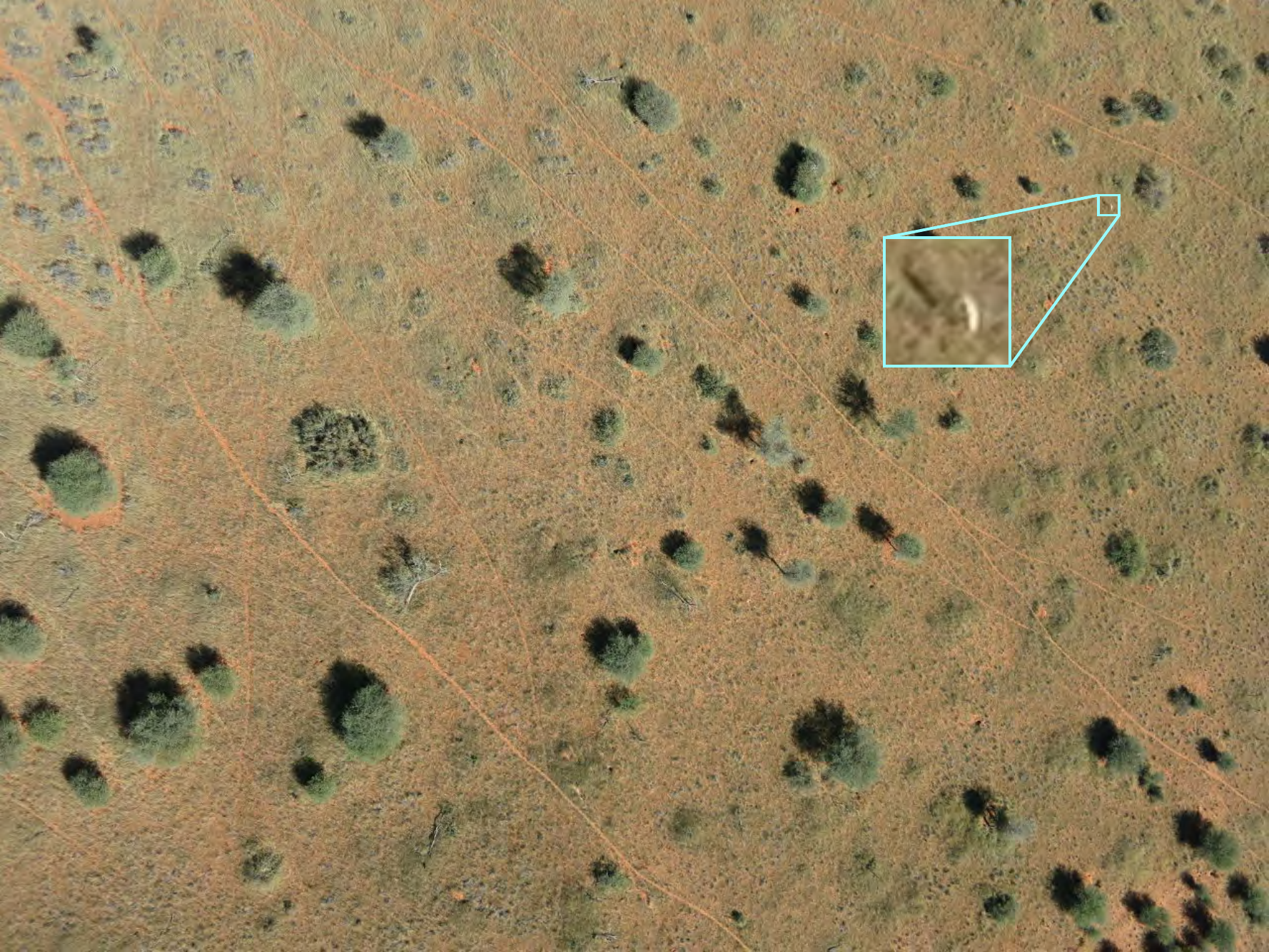}
\caption{Full extents of a sample image. The single animal present in this scene (enlarged in the red bounding box) is difficult to locate, both for humans and machines.}
\label{fig:needleInHaystack}
\end{figure}

Game reserves and national parks are typically concentrated areas, sometimes even confined,  that contain wildlife in numbers above average, compared to the surroundings. In the case of the Kuzikus reserve addressed in this work, the estimated 3000 large animals live in an only moderately large reserve area. However, even under these conditions animals are a very rare sight, which makes livestock censuses a particularly arduous task. This is also the case of the dataset considered in this work, and locating an animal in UAV images corresponds to finding the needle in a haystack (Figure \ref{fig:needleInHaystack}). As a consequence, UAV datasets can rapidly grow in size to dimensions that considerably hamper the performance of detectors.

\begin{figure}
\centering
\begin{subfigure}[b]{0.15\textwidth}
\includegraphics[width=\textwidth]{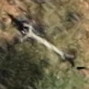}
\caption{}
\label{fig:appearanceConfusion_a1}
\end{subfigure}
\begin{subfigure}[b]{0.15\textwidth}
\includegraphics[width=\textwidth]{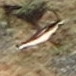}
\caption{}
\label{fig:appearanceConfusion_a2}
\end{subfigure}
\vrule
\begin{subfigure}[b]{0.15\textwidth}
\includegraphics[width=\textwidth]{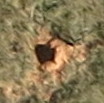}
\caption{}
\label{fig:appearanceConfusion_b1}
\end{subfigure}
\begin{subfigure}[b]{0.15\textwidth}
\includegraphics[width=\textwidth]{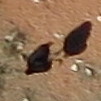}
\caption{}
\label{fig:appearanceConfusion_b2}
\end{subfigure}
\caption{Examples of animal-to-background confusion: dead tree trunks (\ref{fig:appearanceConfusion_a1}) can easily be mistaken for ungulates (\ref{fig:appearanceConfusion_a2}); shadows of dirt mounds (\ref{fig:appearanceConfusion_b1}) may look like ostriches (\ref{fig:appearanceConfusion_b2}).}
\label{fig:appearanceConfusion}
\end{figure}

The reason for performance degradation in large area sizes can be boiled down to two main problems:

\begin{itemize}
\item \textit{(i.)} Increased background heterogeneity. In heterogeneous landscapes such as the African Savanna, an increase in study area size will inevitably include substantially more landscape types, and hence more background variation. Human interpreters are aware of landscape variations and these usually do not cause any issue, but \tc{an animal detector}{a detection algorithm} that has been trained on one set of grounds is likely to fail on areas unseen.

\item \textit{(ii.)} Amplified coupling between the few animals and the background. This refers to the appearance heterogeneity within and homogeneity between classes. Animals themselves have a very diverse set of appearances due to different species, sizes, fur variations, and more. Detectors thus need to be able to learn all these variations as well as possible. At the same time, the expansion to more areas also means that a higher number of background objects looking like animals are included that may confuse a detector. Examples for such similar background objects are tree trunks and dirt mounds, as shown in Figure \ref{fig:appearanceConfusion}. A model that is trained to distinguish between individual animals is thus likely to mistake these background objects, and provide false alarms.
\end{itemize}

All these effects have the consequence that extrapolating results from a small to a big area is not going to yield trustworthy results, and models trained on a small subset and then evaluated on larger areas will not perform satisfactorily.

\section{Addressing realistic and imbalanced datasets}

In this section we discuss steps required to perform and improve semi-automated animal censuses on a realistic dataset. We begin by reviewing the working principle of CNNs in Section~\ref{sec:cnnTheory}. In Section~\ref{sec:cnnTraining}, we introduce a series of practices applicable during the training process of a CNN to make it learn all background variations \tc{without forgetting}{while simultaneously learning} the appearance of the small number of animals. Section~\ref{sec:evalProtocol} then defines the evaluation protocol we used.

\subsection{Working principle of Convolutional Neural Networks}
\label{sec:cnnTheory}

\tc{}{In this section we briefly present the concepts behind CNNs, but without ambitions to be complete. For an in-depth explanation, we refer to the comprehensive work of~\citep{Goodfellow-et-al-2016}.} 
CNNs are the workhorse of deep neural network methods in computer vision\footnote{This is a very rough and non-technical introduction to CNNs. The interested reader will find a more complete overview in the book by Ian Goodfellow and colleagues~\citep{Goodfellow-et-al-2016}.}. Like other deep learning methods, CNNs allow to learn a set of hierarchical features that have been optimized for a given task (e.g. classification, regression, etc.). Their particularity consists in profiting from the inherent structure of images by extracting the features locally. Instead of using the whole image to compute a single feature, the same local feature is extracted in each image location (i.e each pixel), based only on  the values from a few neighboring pixels. In particular, a convolution operation is used to extract the learned features, which generally considers a square neighborhood and captures a pattern of interest (e.g. a vertical green to brown transition) that is learned automatically. This means that, at each image location, we compute the scalar product between the square local neighborhood in the image and the pattern of interest, also called the \emph{``filter''}. This \tc{}{set of local} multiplications is called a \emph{``convolution''.} Convolving the filter at all locations in the image produces a so-called ``activation map'' that contains high values in the regions of the image where the pattern is more present.

After having extracted multiple such activation maps, this stack of maps (also called ``tensor'') can be treated as a new input image, from which new features can be extracted (e.g. a combination of green to brown transitions that indicate the presence of a tree in the image).

Since a composition of linear operators is also a linear operator, there is not much value added to composing a series of convolutions, other than it being equivalent to a single convolution with a larger filter (e.g. composing two convolutions with $3\times 3$ filters can be equivalent to a single $5 \times 5$ convolution). To allow for richer, non-linear, relationships between the input and the output, a simple non-linear function (e.g. a rectifier, sigmoid, etc.) is interposed between each pair of convolution operators. In addition to this, it is common to perform downsampling after some of the convolutions, which reduces the size of the tensor of activations and, therefore, increases the receptive field, i.e. the effective area in the image that passes information to each local feature in the following layer.

CNN filters are generally initialized with random values or with values learned on some previous problem. In both cases, the filters can be improved for some particular task by defining a differentiable cost function that measures the goodness of the current solution. In our case, we employ the so-called ``cross-entropy loss'' that is standard for classification tasks:

\begin{equation}
  \mathcal{L}(y,\hat{y}) = - \sum_c w_c y_{c}\log(\hat{y_c})\label{eq:loss}
\end{equation}

In this loss, the prediction $\hat{y}$ of the CNN is compared to the true label $y$ for every class $c$. Also note the weight $w_c$ that amplifies or dampens the effect per class in comparison to the others\tc{}{, achieved by altering the learning rate itself depending on the true class for the respective sample}.

\tc{}{
For a given learnable parameter $\theta$ (\emph{e.g.} an element in one of the convolutional filters), the gradient of the loss with respect to $\theta$ is computed. While this can be done directly for the last layer, the chain rule has to be applied for all hidden layers before, as their gradients depend on the gradients of the following layers. This update procedure is known as backpropagation and requires all operations used in the CNN to be differentiable. After computing the gradients with respect to each parameter, the parameters are updated with the following rule:
\begin{equation}
\theta \leftarrow \theta - \eta \frac{\partial \mathcal{L}(y,\hat{y})}{\partial \theta},
\end{equation}
where $\eta$ is the learning rate.
}

Evaluating such function allows to improve the filters by gradient descent~\citep{Goodfellow-et-al-2016}. By applying this operation for multiple iterations, often several thousands, we can obtain convolutional filters for each of the layers that have been optimized for solving our problem. In sum, CNNs thus are not only able to learn a classifier, but also a series of expressive filters, produced by the convolutional operators, which allows them to yield superior performance in image-based tasks.

\subsection{Training deep object detectors on imbalanced datasets}
\label{sec:cnnTraining}

CNNs are commonly trained on large datasets containing thousands of examples for each class, and all classes often occur in similar percentages. For instance, the ImageNet dataset~\citep{russakovsky2015imagenet} contains thousands of categories and more than 14 million images in total\footnote{as of March 2018; \url{http://www.image-net.org/}}. In our case, the overall amount of data is substantially smaller, 
\tc{
but the dataset poses two particularly delicate problems that are not found in datasets like ImageNet: \textit{(a.)} the background class substantially dominates over the animal class in terms of the number of examples; \textit{(b.)} the small number of animal examples is most likely not sufficient for a complex model like a CNN to learn all required appearance properties of the animals. 
}{
but poses the additional challenge of data imbalance, with the background class being overwhelmingly larger than the positive class, both in terms of quantity and sample complexity.
}
As a result, a CNN trained \tc{na\"{i}vely}{without specific alterations in the training plan} on such a naturally unbalanced dataset will be flooded by background examples, and will inevitably miss most animals.

In order to account for these problems, a CNN must be trained in such a way that it learns all possible variations in the background, while \tc{not forgetting}{also addressing} the appearance of the few foreground instances it has seen. The entire process corresponds to finding an equilibrium between two extremes (not detecting the animals at all and firing false alarms everywhere), which can be very delicate. In the following, we present a series of recommendations applicable during CNN training that will steer the model in the right direction. The effects of each individual procedure will be assessed in an ablation study, whose results are presented in Section~\ref{sec:ablationStudies}.

\subsubsection{Addressing the class imbalance}\label{ssec:w}

An initial solution to overcome the imbalance problem might be to artificially balance the dataset by oversampling, i.e. repeating each animal instance to match the total number of background locations. While this has been shown to work well for other tasks~\citep{buda2017systematic}, we empirically found it to cause the CNN to overfit to the small number of animal instances present in the training set. The inverse, i.e., reducing the number of background samples to match the number of animals (``undersampling'') has a similar effect in that the model fails to learn the variability of the background, which leads the model to misdetect everything that looks even remotely similar to animals.

We instead apply conventional class weights to reduce the impact of the background class. In practice, we set the class weights $w_c$ in Eq.~\eqref{eq:loss} to different values, to ensure that an error in the animal class counts much more than one in the background class. Exact values depend on the data value range, the training schedule and the amount of data the CNN is trained on and may thus vary from problem to problem. Nevertheless we got satisfactory result by weighting classes according to the inverse of the frequency in which they occurred in the training set.

\subsubsection{Making the model learn all background variations}

Large natural datasets contain a lot of variations in appearance for both the animals and background classes. This causes an increased number of false alarms. To avoid such confusion, the CNN needs to be trained on as much of the background area as possible. Below we present techniques that can employed to achieve such goal.

We start with the observation that, thanks to their convolutional nature, CNNs can process inputs of arbitrary spatial dimensions, up to the limits of the available memory. This has the effect that CNNs can be trained not only on single patches around specific locations, but on larger images. In such a setting, evaluating on a larger image will not yield a single label, but a spatial grid of activations. An advantage of proceeding this way is that neighboring locations partially share activations of overlapping filters, and this way a much higher data throughput per training iteration can be achieved, as less redundant filter activations need to be stored for the backward pass. \tc{Enlarging the training images therefore has a similar effect to increasing the number of single-label patches, while being more memory-efficient and reducing the undesired lack of spatial context at the image borders.}{Using larger training images therefore has a similar effect to increasing the number of patches, while being more memory-efficient. Moreover, it allows the model to make full use of its receptive field, thus capturing as much context as possible and reducing the overall impact of border effects. Nevertheless, this size is limited by amount of memory available on the GPU for storing the intermediate activations.} We achieved satisfactory results by training models with image patches of size $\mathrm{512\times512}$ pixels, cropped from the larger images at semi-random locations: if a training image contains animals in the ground truth, we crop the patch so that it encompasses at least one animal instance. In images without an animal, the location is randomly chosen at every iteration to maximize the learned background diversity.

\subsubsection{Curriculum learning as a starting boost}

CNN training corresponds to a series of epochs in which portions of the training dataset are fed to the network in random order. Consequently, it is an iterative process. Since the training samples picked during each iteration might only reflect parts of the full dataset distribution, the ordering can be very decisive in the quality of the final model. Although the model parameters are constantly being adjusted, chances are that it ``forgets'' the representation of certain instances once the learning process shifts to other parts of the dataset. In our case, this risk is related to the class imbalance: if the CNN is trained on the full dataset from the start, the background class might drown the signal emerging from the animals, despite the class weights (see Section~\ref{ssec:w}). We propose to countersteer this effect based on the concept of curriculum learning~\citep{bengio_curriculum}, where a model is adaptively trained with different portions of the dataset. While curriculum learning is commonly applied in cases where datasets show gradually increasing complexity, we employ it to force the CNN to learn a more balanced representation of both animals and background in the beginning. To this end, we start training the model on a subset of the training data that has been artificially restricted to image patches that contain at least one animal. This way, the CNN has the chance of learning how animals should be represented, while having only an approximate knowledge of the variability of the background. We use this schedule for five epochs, and then switch over to the full dataset, where image patches might not always have an animal. After this switch, the model is confronted to the entire variability of the background class.

\subsubsection{Rotational augmentation}
\label{sec:rotAug_intro}
\tc{}{One practice commonly employed during the training of CNNs is \emph{data augmentation}. This consists of creating artificial training samples by altering the existing one in realistic ways. A common augmentation strategy is to rotate images randomly during training. Since our UAV dataset has a perpendicular top-down perspective, rotating the image has a similar effect to acquiring the ground with the UAV at a different angle. In theory, applying rotations at random to images should make the detector more robust to grounds and animals occurring at different orientations in the field. In other words, rotational augmentation allows us to explore a larger set of possible viewing angle versus animals configurations.}

\tc{}{However, a critical side-effect of applying extensive augmentation is that it significantly changes the dataset seen by the model itself. For example, if the training images are rotated at random with a probability of 50\%, the model sees up to 1.5 times as many data points as before. In the case of rotations, where usually every pixel is altered significantly, this might lead to too strong perturbations. Especially at the beginning of training, a model is supposed to get an overall glimpse of the statistical distribution of animals and background in the dataset. Instead, we limit data augmentation to random horizontal and vertical flips of the images with a 50\% probability each during the first 300 epochs. Only afterwards, we introduce random rotations with a 75\% probability and train the model for another 100 epochs with a at this stage slightly reduced learning rate. We limit rotation angles to multiples of 90 degrees to avoid spatial shifts of the (reduced size) ground truth due to nearest-neighbor interpolation. More advanced augmentation techniques, such as synthetic data generation using conditioned generative adversarial networks~\citep{shrivastava2017learning,radford2015unsupervised} are exciting  perspectives of our system that might be considered in the future.}

\subsubsection{Hard negative mining}
\label{sec:hardNegativeMining}

The general idea of class weights and curriculum learning explained above is to prevent one class from dominating over the other. This is especially important at the beginning of the training procedure, where the model parameters are not yet optimized enough to the problem at hand. However, after a sufficient number of iterations, a state will usually be reached where the model performs fairly well in finding the animals, but might not reach its maximum precision. At this stage, the model performance can be further boosted to reduce the number of false alarms by shifting from treating all data points equally to focusing on the errors. We propose to do so by means of hard negative mining~\citep{malisiewicz2011ensemble,Shrivastava_2016_CVPR}. This technique essentially gives special treatment to the most severe mistakes, which in our case are the false positives scored with the highest confidence. In practice, we train the CNN regularly for 80 epochs and then employ hard negative mining by locating the four background samples in the $\mathrm{512\times512}$ grid that have been scored with the highest confidence, and assigning them a dedicated weight that is \sfrac{1}{4} of the animal class weight. We do not assign more weight to prevent the model from forgetting the animal appearances, as our ultimate goal is to maintain a high recall (see Section~\ref{sec:evalProtocol}).

\subsubsection{Introduction of the border class}
\label{sec:borderClass}

\begin{figure}
\centering
\begin{subfigure}[b]{0.3\textwidth}
\includegraphics[width=\textwidth]{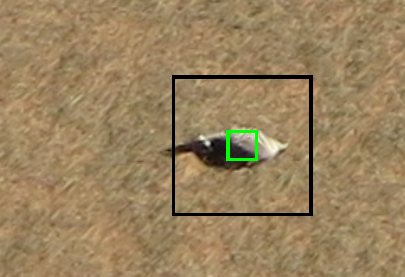}
\caption{Receptive field on an animal}
\label{fig:CNNborderClass_a}
\end{subfigure}
\begin{subfigure}[b]{0.3\textwidth}
\includegraphics[width=\textwidth]{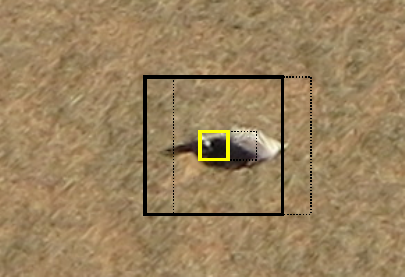}
\caption{Receptive field at the border}
\label{fig:CNNborderClass_b}
\end{subfigure}
\caption{A CNN takes into account multiple neighbors of a specific location (``receptive field'') to determine the likelihood of the center location being an animal (left). To avoid confusion of animals not in the center, but still in the receptive field (right), we introduce a border class around a true location.}
\label{fig:CNNborderClass}
\end{figure}

A slightly less obvious problem of training a model on larger patches instead of single examples is the effect of \emph{``spillage''} along the spatial boundaries between locations of different classes. CNNs classify locations (in our case, into ``animal'' vs. ``background'') by taking into account the location's neighboring pixels via the model's receptive field. For instance, in Figure \ref{fig:CNNborderClass_a}, the location of interest (in green) is predicted based on the CNN's receptive field size (in black). While this works as intended at the center of an animal, it might fail in its surroundings (Figure \ref{fig:CNNborderClass_b}): in this case, the final label should be ``background'' (yellow square) to avoid multiple predictions of the same object. However, the CNN's receptive field still includes a major portion of the animal itself. Training using the latter case risks to produce the undesirable effect of the CNN learning that patches influenced by animals through the receptive field belong to the background class. Ignoring these locations in turn (setting the weights to zero) will give the model too much freedom in exact positioning, which might lead to false alarms in the surroundings of the center location. This is particularly problematic with animals standing close-by, as they cannot be properly separated anymore.

An initial solution to this problem is to decrease the CNN's receptive field in order not to include nearby animals. This however has the undesirable effect of providing a too fine grid of label predictions. As a consequence, an animal may suddenly encompass multiple locations, causing multiple neighboring false alarms that need to be suppressed~\citep{brutalMetalNonsense}. Furthermore, a too small receptive field will make the CNN fail in capturing the full appearance of the animal.

\begin{figure}
\centering
\begin{subfigure}[b]{0.25\textwidth}
\includegraphics[width=\textwidth]{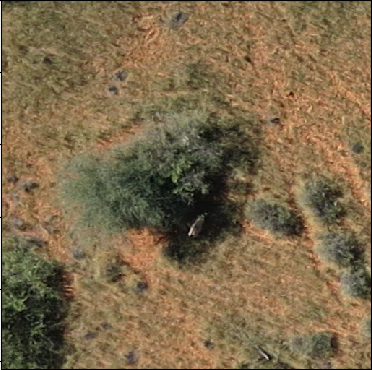}
\caption{Image patch}
\label{fig:CNNborderClass_example_input}
\end{subfigure}
\begin{subfigure}[b]{0.25\textwidth}
\includegraphics[width=\textwidth]{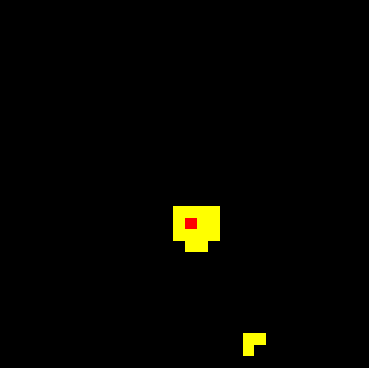}
\caption{Model prediction}
\label{fig:CNNborderClass_example_prediction}
\end{subfigure}
\begin{subfigure}[b]{0.25\textwidth}
\includegraphics[width=\textwidth]{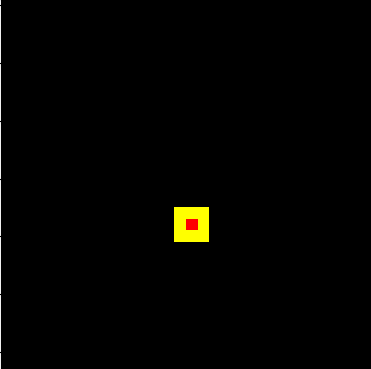}
\caption{Ground truth}
\label{fig:CNNborderClass_example_groundTruth}
\end{subfigure}
\caption{Example image patch (left), its prediction grid by the CNN (middle), and the corresponding ground truth (right). As can be seen, the CNN learns to distinguish between background (black), animals (red) and the border class (yellow) fairly well.}
\label{fig:CNNborderClass_example}
\end{figure}

We propose to address this issue by instead including a dedicated ``border class'', which corresponds to locations that technically belong to the background, but still include portions of an animal. The effect of a border class is that the CNN learns to treat spatial transition areas separately and predict them accordingly. At test time, locations predicted as ``border'' by the CNN can simply be discarded, leaving the predicted center location as the only detection. Multiple ways of including a border class are possible and depend on the sizes of the animals and CNN prediction grids. In our case, we obtained satisfying results by assigning the eight neighboring pixels around an animal center to the border class (see Figure~\ref{fig:CNNborderClass_example}).

\subsection{Census-oriented evaluation protocol}
\label{sec:evalProtocol}

Although many census-based studies include evaluations using common metrics on a held-out test set, they do not explain the exact criteria required by a detection to be counted as a correct match. While this is well established for classification (a data point is hard-assigned to a class), it is less clear for object detection. For instance, if the main target is not to retrieve the exact position of animals, the question of how far away predicted animals can be from the ground truth locations arises. \tc{}{Traditional computer vision measures, such as the Intersection-over-Union~\citep{everingham2010pascal}, are intended for spatially precise object detections and would penalize slight spatial deviations too harshly for our application.} Also, the handling of multiple predictions for the same \tc{}{animal} is generally not explicited. This is especially important in the case of multiple animals standing close-by each other.

\begin{figure}
\centering
\begin{subfigure}[b]{0.22\textwidth}
\includegraphics[width=\textwidth]{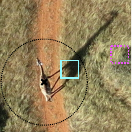}
\caption{}
\label{fig:censusProtocol_singleAnimal}
\end{subfigure}
\begin{subfigure}[b]{0.22\textwidth}
\includegraphics[width=\textwidth]{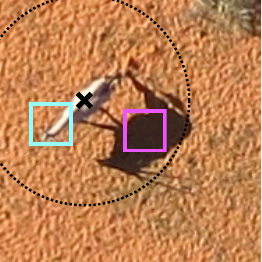}
\caption{}
\label{fig:censusProtocol_singleAnimal_2}
\end{subfigure}
\begin{subfigure}[b]{0.22\textwidth}
\includegraphics[width=\textwidth]{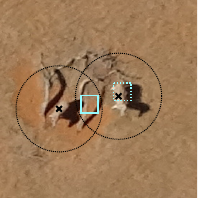}
\caption{}
\label{fig:censusProtocol_multiAnimal}
\end{subfigure}
\begin{subfigure}[b]{0.22\textwidth}
\includegraphics[width=\textwidth]{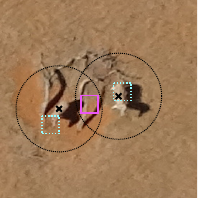}
\caption{}
\label{fig:censusProtocol_multiAnimal_2}
\end{subfigure}
\caption{Example cases of the census-oriented evaluation protocol. Predictions outside a certain distance range (\ref{fig:censusProtocol_singleAnimal}) as well as multiple predictions of the same animal (\ref{fig:censusProtocol_singleAnimal_2}) are rejected (magenta). In the case of a prediction matching more than one animal, it is counted as a true positive (cyan) if at least one animal is not covered by another prediction (\ref{fig:censusProtocol_multiAnimal}), or \tc{rejected}{marked} as a false positive if all are (\ref{fig:censusProtocol_multiAnimal_2}).}
\label{fig:censusProtocol_examples}
\end{figure}

To account for these problems, we propose an evaluation protocol that reflects the final objective: to provide animals abundance, whereas the exact pixel location is only secondary. This allows us to judge the quality of each prediction according to the following set of rules:

\begin{itemize}
\item Predictions may be candidates for correct detections if their position falls within a (circular) range of reasonable size around each ground truth's center location.
\item A ground truth object is counted as correctly identified if at least one prediction falls within the distance range.
\item $n$ predictions inside the distance range are counted as one true positive, and $(n-1)$ false alarms.
\item If a prediction lies within the distance range of more than one ground truth location, it may be counted as at most one true positive, given that there is at least one ground truth object that has no other prediction in its range. If all involved ground truth objects are already predicted by other predictions, the current one is discarded as a false positive.
\end{itemize}

Consider the examples in Figure \ref{fig:censusProtocol_examples}. In situation \ref{fig:censusProtocol_singleAnimal}, there are two detections in vicinity of a single giraffe, but both of them miss it spatially by quite a margin. However, the one closer to the animal (cyan) actually detects its shadow, and should thus be counted as a true positive. The other (magenta dashed) lies outside the detection range and is thus \tc{rejected as}{considered} a false positive. Situation \ref{fig:censusProtocol_singleAnimal_2} shows one animal surrounded by multiple predictions, all of which lie inside the allowed distance range. Since our main target is providing animal counts, only one prediction (generally the closest) should be counted as a correct detection (cyan), and the others dismissed as false alarms (magenta). However, such a rule cannot always be applied straight away, as situation \ref{fig:censusProtocol_multiAnimal} shows: here, two animals are surrounded by multiple detections, out of which one falls directly in between the two (cyan). In this case, the prediction in-between could be counted as a false alarm considering that the animal to the right is already predicted by another detection (cyan dashed). However, as the other animal has no other detection in its vicinity, both predictions should be treated as correct in this case. If, however, all involved animals are already covered by an equivalent number of predictions (case \ref{fig:censusProtocol_multiAnimal_2}), any superfluous prediction, such as the one in the middle (magenta), is \tc{rejected}{marked} as a false positive. In all cases, animals completely missed by the detector are counted as false negatives.

The effect of such a procedure is that the number of true positives cannot exceed the actual number of animals, which corresponds to a realistic census scenario. A hypothetical ideal detector will reach a perfect score if the number of predictions matches the number of actual animals, and if they are reasonably close to the ground truth positions. The only parameter to be set manually is the distance range around ground truth objects, and we will evaluate effects of different ranges in Section~\ref{sec:AblationStudy_distanceThresholds}. The protocol is nevertheless robust enough even at less reasonable ranges: if the distance range is set too small, the penalization is increased and only detectors that provide a good enough positional accuracy will yield good scores. In turn, an unreasonably large distance range will include more potential predictions further apart, but multiple detections are still dismissed as false alarms. In all cases, the protocol will assign the best scores if the number of predictions matches the number of animals present in the scene, which is the ultimate goal of census campaigns.


\section{Experiments and results}

In this section we present the effects of the training practices described above. We start by describing the models we used (baseline and CNN) in Section~\ref{sec:modelsUsed}. The CNN training recommendations\tc{}{, including rotational augmentation,} are then put to test in a dedicated number of ablation studies (Section~\ref{sec:ablationStudies}). A third chapter addresses the effects of choosing different evaluation distance thresholds (Section~\ref{sec:AblationStudy_distanceThresholds}). Finally, Section~\ref{sec:resultsFullDataset} presents the results obtained on the full dataset based on the best baseline and CNN architectures, respectively.

\subsection{Models setup}
\label{sec:modelsUsed}

\subsubsection{CNN with proposed training schedule}
\label{sec:cnnTrainingSchedule}

\begin{figure}
\includegraphics[width=\textwidth]{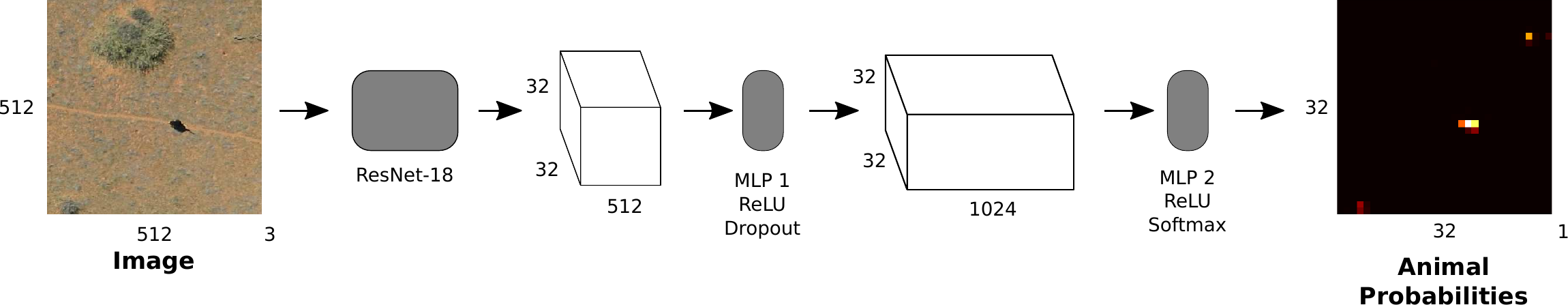}
\caption{Working principle of the CNwN-based animal detector. We train the model on images of size $\mathrm{512\times512}$, which results in a grid of $\mathrm{32\times32}$ locations with animal probability scores for each location.}
\label{fig:CNN_architecture}
\end{figure}

We base all experiments on a single CNN architecture, a simplified version of \citep{brutalMetalNonsense} that performs detection, but no regression of bounding boxes around the animals. The overall architecture is shown in Figure~\ref{fig:CNN_architecture}. The base feature extractor consists of the first convolutional and four subsequent residual blocks of an instance of ResNet-18~\citep{he2016deep}. This model had been pre-trained on the ImageNet classification challenge~\citep{russakovsky2015imagenet}, i.e. has been trained on the task of classifying visual classes that are not specific to wildlife monitoring. 
To enforce the model to be specific to our problem, we adapt it by adding two one-by-one convolutional blocks or Multi-Layer Perceptrons (MLPs) with nonlinear activations (ReLU), dropout regularization~\citep{srivastava2014dropout}, and softmax activation at the end. This way of proceeding, also called \emph{fine-tuning}, is a very common strategy used to adapt these models to specific applications, since generic deep learning architectures (such as ResNet) are very good at extracting relevant generic local features in the earlier layers and only need to be made specific to the problem at hand in the deeper layers~\citep{castelluccio2015arxiv}.\tc{\newline At test time, we slide the model over each whole image and retrieve a prediction grid coarser in resolution, with confidence scores for every grid cell and hence location in the image to contain an animal. \tc{}{In detail, we divide each $4000\times3000$ image into $8 \times 6$ sub-patches of size $512\times512$ and evaluate each patch individually, yielding a prediction grid of $32 \times 32$ each. Note that the patches do overlap at their borders to some extent; in those areas we eventually average the obtained class probabilities when stitching together the patches. Eventually, this yields a $188 \times 250$ prediction grid for an entire UAV image.}}{}

To assess the training recommendations presented in Section~\ref{sec:cnnTraining}, we trained a series of CNNs in a held-out fashion: for each model, we enable all but one of the five \tc{}{main} recommendations presented above (and summarized in Table~\ref{tab:CNNmodels}). The specific properties and parameters of the recommendations are given in Table~\ref{tab:CNNparameters}. The final model (referred to as ``Full Model'') had all recommendations enabled. \tc{}{We separately analyzed the effect of rotational augmentation in a dedicated ablation study (Section~\ref{sec:AblationStudy_rotAug}).} All models were trained using the Adam optimizer~\citep{kingma2014adam}, with momentum of 0.9 and a learning rate gradually decreasing from $10^{-4}$ (first 5 epochs) to $10^{-5}$ (next 5 epochs), to $10^{-6}$ (next 100 epochs), and then again to $10^{-7}$ for the rest (until final epoch 400). \tc{}{Weight decay helped reducing overfitting effects and was employed with a factor of $10^{-3}$ for the first five epochs, and $10^{-4}$ for the rest.} \tc{Besides semi-random patch cropping, additional data augmentation was employed in the form of random mirroring along both image axes, each with a probability of 50\%. \tc{}{More advanced augmentation techniques, such as synthetic data generation using conditioned generative adversarial networks~\citep{shrivastava2017learning,radford2015unsupervised} are conceivable means for CNN training enhancements, but we leave them for further experiments.}}{}

\tc{}{At test time, we slid the model over each whole image and retrieved a prediction grid coarser in resolution, with confidence scores for every grid cell (and hence location) in the image to contain an animal. In detail, we divided each $4000\times3000$ image into $8 \times 6$ sub-patches of size $512\times512$ and evaluated each patch individually, which yielded a prediction grid of $32 \times 32$. Note that the patches did overlap at their borders to some extent; in those areas we eventually averaged the obtained class probabilities when stitching together the patches. Eventually, this yielded a $188 \times 250$ prediction grid for an entire UAV image.}

\tc{}{All CNN-based models were implemented in PyTorch\footnote{\url{http://pytorch.org/}}. Training time per model took approximatively four days on a Linux workstation with an Intel Xeon CPU and an NVIDIA GeForce GTX 1080Ti graphics card.}

\begin{table}
\centering
\begin{tabular}{c | c c c c c | c}
 & Class & Full & Curr. & Hard & Border & Rot.\\
Model & Weights & Dataset & Learning & Negatives & Class & Augm.\\
\hline
CNN 1 & & \checkmark & \checkmark & \checkmark & \checkmark &\\
CNN 2 & \checkmark & & \checkmark & \checkmark & \checkmark &\\
CNN 3 & \checkmark & \checkmark & & \checkmark & \checkmark &\\
CNN 4 & \checkmark & \checkmark & \checkmark & & \checkmark &\\
CNN 5 & \checkmark & \checkmark & \checkmark & \checkmark & &\\
\tc{}{CNN 6} & \checkmark & \checkmark & \checkmark & \checkmark & \checkmark &\\
\hline
Full Model & \checkmark & \checkmark & \checkmark & \checkmark & \checkmark & \checkmark\\
\end{tabular}
\caption{The CNN models trained with different recommendations: for the ablation study, five models were trained with one of the recommendations held out (top rows); the final model (bottom row) used all recommendations. ``Full Dataset'' denotes models were trained on $\mathrm{512\times512}$ patches both with and without animals; CNN 2 only saw patches that contained at least one animal. \tc{}{The final model (``Full Model'') corresponds to CNN 6, further fine-tuned with rotational augmentation from epoch 301 to 400.}}
\label{tab:CNNmodels}
\end{table}

\begin{table}
\centering
\begin{tabular}{l | l}
Recommendation & Parameters\\
\hline
Class weights & $1.0$ (animals), $\sfrac{1}{80}$ (background), $\sfrac{1}{8}$ (border) \\
Curriculum learning & 5 epochs only on images with animals\\
Hard negative mining & weight 0.5 for 4 hard negatives (from epoch 80)\\
Border class & over 8-neighborhood around animal \\
\tc{}{Rot. augmentation} & \tc{}{75\% random $90^{\circ}$ stops [${\pm}270^{\circ}$] (from epoch 301)}
\end{tabular}
\caption{Parameters for the recommendations (where applicable), used to train the full CNN.}
\label{tab:CNNparameters}
\end{table}

\subsubsection{Baseline model}

As a baseline, we employ the current state-of-the-art on the Kuzikus dataset presented in \citet{rey2017detecting}. This model relies on a pre-selection of candidate locations (``object proposals''), which are then classified in a second step. We use the same object proposals and feature extraction methods proposed in the original work, but replace the original exemplar SVM with a random forest classifier, which is a solid baseline in remote sensing, providing nonlinear response at no extra computational expense~\citep{Pel16}. We found the random forest to provide a similar performance as reported in \citet{rey2017detecting}, which can also be seen in the precision-recall curves in Figure \ref{fig:PRcurve_final_testSet}.

\subsection{Ablation studies}
\label{sec:ablationStudies}

In this section, we first evaluate the impact of the five \tc{}{basic} recommendations for effective CNN training provided in Section~\ref{sec:cnnTraining}\tc{. Later on,}{, followed by a study on rotational augmentation. Finally,} we also assess the role played by the distance threshold in our assessment procedure.

\subsubsection{Effects of the recommended CNN training recommendations}

\begin{figure}
\centering
\includegraphics[width=12cm]{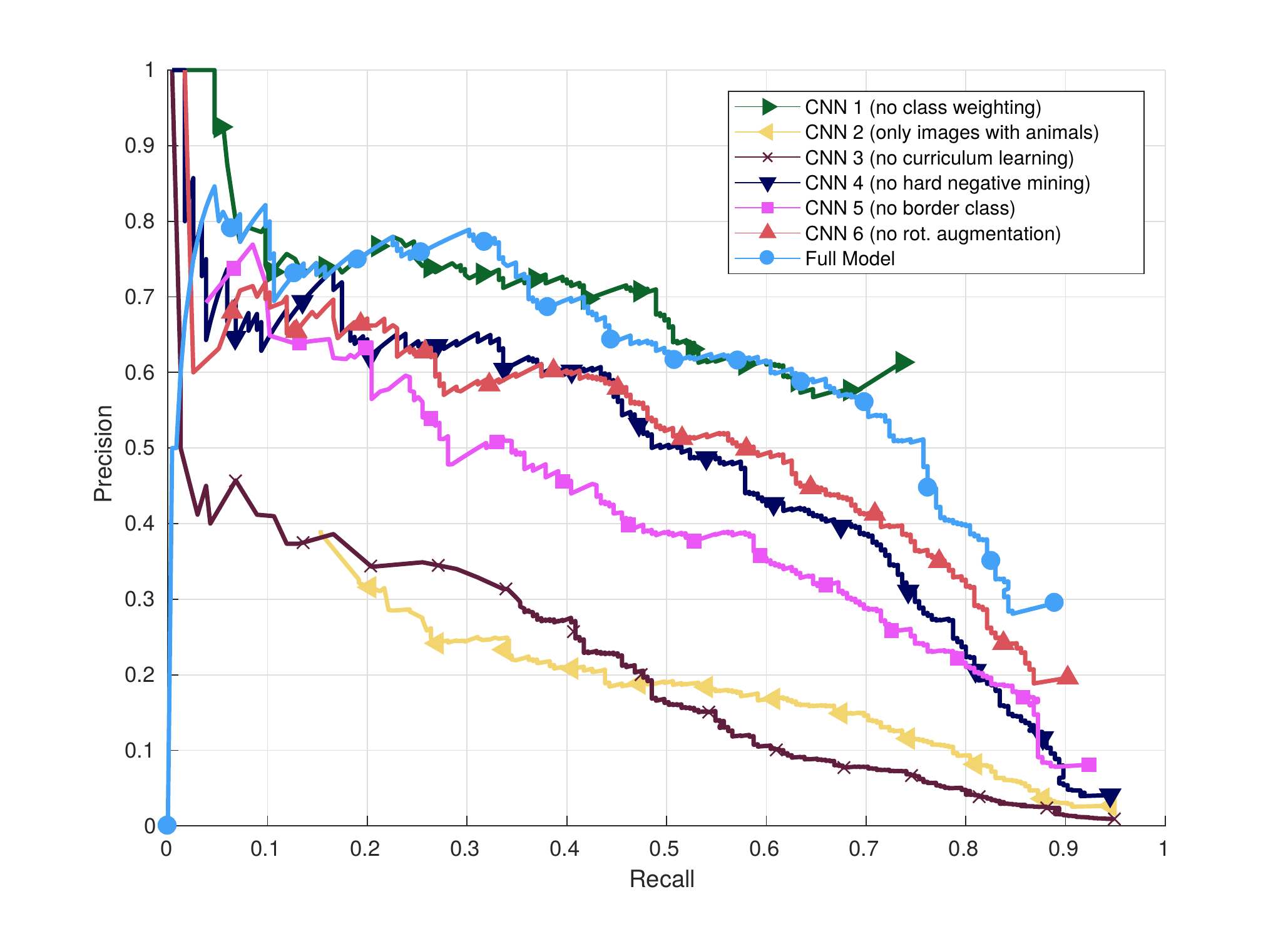}
\caption{Precision-recall curves for the different CNNs on the test set\tc{}{, based on the CNN confidence scores}. All recommended training recommendations provide an increase in precision at given recall values, but the best model emerges when all of them are combined together\tc{}{, and when the model is further fine-tuned using 90 degree-stop rotational data augmentation}.}
\label{fig:ablationStudy_CNN_prCurve}
\end{figure}

Figure \ref{fig:ablationStudy_CNN_prCurve} illustrates the results obtained by the \tc{six }{}CNNs described in Section~\ref{sec:cnnTrainingSchedule} on the complete test set. As can be seen, almost every aspect of the proposed CNN training plan in itself provides an improvement in precision, but mostly in regions of high recall (beyond 80\%). The best model is the one trained with all recommendations applied together. The effect of each individual recommendation does not seem to scale linearly, but rather to depend on the enabling of the other recommendations.

Quite surprisingly, ``CNN 3'', the model trained without curriculum learning (that is, a model trained with the full training set from the beginning) yields the worst precision at high recall values. The CNN 2 model, trained only with the animals containing animals, also performs poorly and only marginally better than CNN 3, therefore implying that only including more background data from the start to train the CNN is not sufficient. Instead, the dominating background class forms a hurdle that can only be overcome by injecting more background patches, intertwined with curriculum learning. This could be explained with the training behavior of CNNs at early epochs: at the beginning of the training, the CNN weights are not yet set to values where their prediction yields a (local) optimum, that is, a low value in the loss function. The model requires several passes over the dataset for the weights to become meaningful for the problem at hand. In an imbalanced setting as mammals detections in the wild, this initial phase appears to be especially critical, as the model needs to learn sufficient data about \emph{both} the background and animals classes. Once this trade-off between the two classes is reached, the models can then be fine-tuned to learn more background variations, thus harnessing all the potential of curriculum learning.

Confusion can further be dampened if the border class is introduced: in fact, the model trained without the border class (``CNN 5'' in Figure~\ref{fig:ablationStudy_CNN_prCurve}) yields the third-lowest performance on the test set and performs especially poorly at high recall rates, for which for every positive example, approximatively 19 false alarms are recorded (5\% precision). We argued above that the border class helps in reducing confusion as the CNN still includes parts of the animals nearby due to its spatial receptive field. Although slightly more CNN parameters need to be learned with every additional class, the results indicate that the border class does indeed guide the CNN to a better solution by distinguishing between animal centers and locations surrounding them.

The model that omitted hard negative mining (``CNN 4'' in Figure~\ref{fig:ablationStudy_CNN_prCurve}) performs similarly to the full model. Such behavior could be expected, given that hard negative mining mostly attempts at boosting the precision at very late training stages. We observe that the technique comes into play at the most critical regions of 90\% recall and more, where it manages to increase the precision from around 5\% (CNN 4) to 20\% (full model). In numbers, this results in a reduction of false alarms from 19 to 4 for every detected animal. While curriculum learning helps in initiating the model in the best way, hard negative mining takes over at the final fine-tuning stage. Clearly, both techniques seem to play a major role in improving the model performance, and thus reducing the amount of time required to manually verify the detections, as we will discuss in Section~\ref{sec:resultsFullDataset}.

Perhaps the most surprising result emerged from the model that was not trained with balancing class weights (``CNN 1'' in Figure~\ref{fig:ablationStudy_CNN_prCurve}). We repeated this schedule several times and observed the same behavior in all runs: the model would predict every location as ``background'' for about 45 to 55 epochs, but would then suddenly raise to a precision of almost 100\% and a recall of around 40\% in the training and validation sets---all within one epoch. The model would then improve on the recall over the remaining epochs. Once the model learns to correctly predict the background class (which is the majority of the samples), the only error signals would emerge from the few animal instances, making the model focus on them. However, the definite cause for the model to jump from one extreme state to another in just one epoch still needs to be resolved. In any case, we trained this model for the full 300 epochs like the others, and although it reaches superior precision for the most part, it fails to exceed in recall over around 74\% of the animals. In other words, this model would never be able to retrieve the remaining 26\% of the animals, no matter the detection threshold selected. In livestock estimations, missing over a quarter of the animals might be fatal. We hypothesize that this corresponds to a substantial overfitting effect to the training set, which means that this model is by far not apt to be used in real census scenarios despite being the one providing the highest precision for smaller recall rates.

\subsection{Rotational augmentation}
\label{sec:AblationStudy_rotAug}

\tc{}{As explained in Section~\ref{sec:rotAug_intro}, we employ random rotations as one of the data augmentation strategies. We decided to assess rotational augmentation in a separate study, as we found it to have remarkably fluctuating influences on the final model performance, depending on the training stage at which rotational augmentation is employed. }
\tc{}{In detail, we experimented with three different training schedules:
\begin{itemize}
\item Training a model with all recommendations and rotational augmentation from the beginning
\item Training with all recommendations, but enabling rotational augmentation from epoch 150 
\item Training with all recommendations, but enabling rotational augmentation from epoch 300 for another 100 epochs
\end{itemize}}

\begin{figure}[!t]
\centering
\includegraphics[width=12cm]{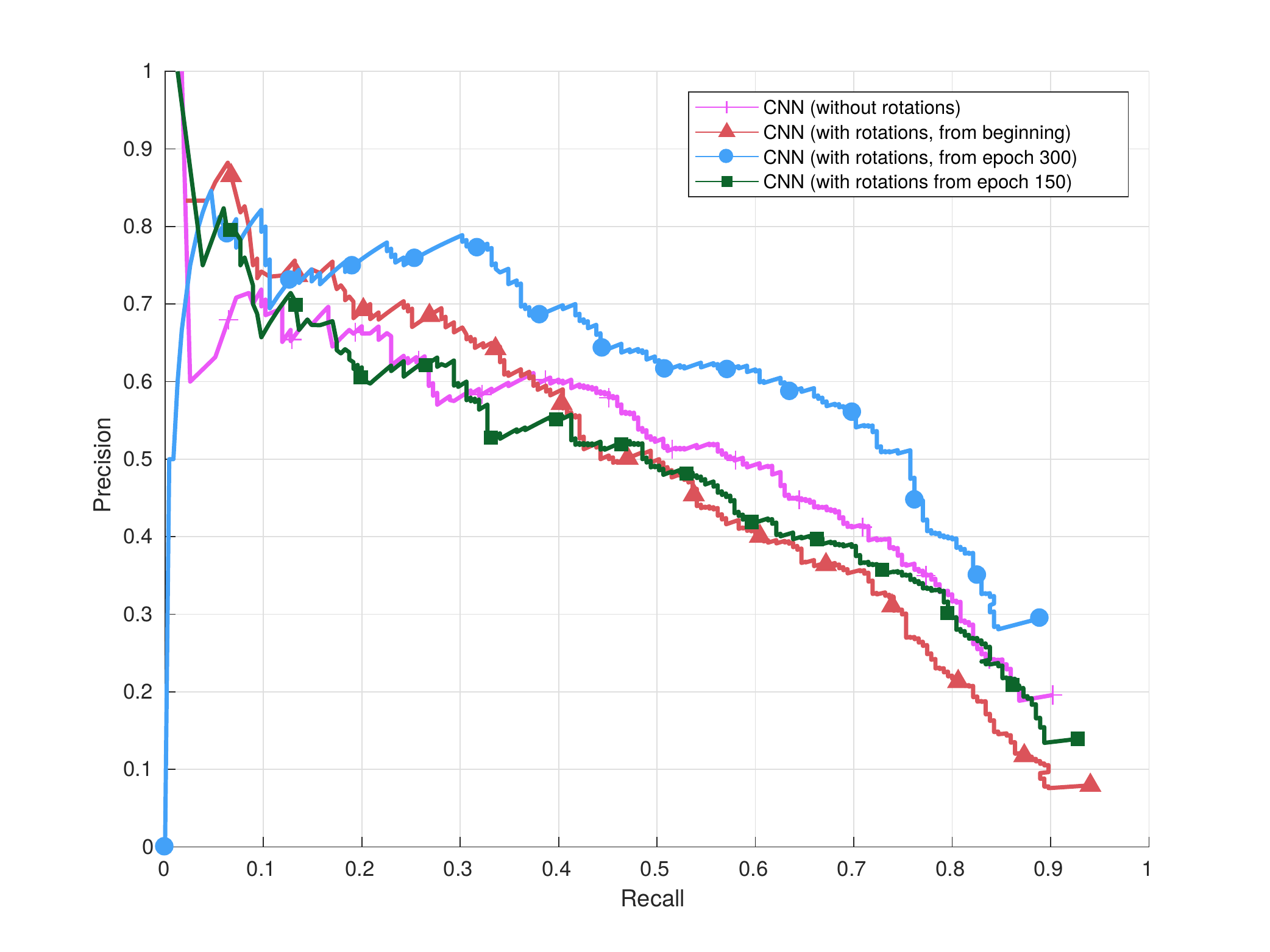}
\caption{\tc{}{Precision-Recall curves on the test set for the ablation study on rotational augmentation.}}
\label{fig:PRcurve_rotAug}
\end{figure}

\tc{}{The resulting precision-recall curves for the these models can be seen in Figure~\ref{fig:PRcurve_rotAug}. Compared to the initial model that was trained with all recommendations except for rotations (pink), two opposing effects of rotational augmentation can be observed, depending on the stages it was employed. If rotations are enabled at relatively early phases in the CNN training, such as epoch 150 (green) or even from the start (red), the final accuracy of the model is similar or worsens: for similar recall values, lower precision figures are observed. As it seems, rotations do have an influence significant enough to slightly confuse the detector at these early stages. If, however, the model is only fine-tuned from epoch 300, also with a lower learning rate, we see a substantial improvement in precision, with only a slight sacrifice in overall achievable recall. Based on these findings, we may conclude that rotations are indeed a valuable augmentation strategy, when applied at later training stages and with gentler learning rates.}

\subsection{Evaluation of the distance thresholds}
\label{sec:AblationStudy_distanceThresholds}

In Section~\ref{sec:evalProtocol}, we introduced an evaluation protocol that works by addressing predictions within a certain distance range that has to be manually set. Sensible thresholds depend on the data resolution, target sizes, and positional accuracy of the model predictions. In this section, we assess the impact of different evaluation thresholds on predictions from both the baseline (random forest following the protocol of \citet{rey2017detecting}) and CNN models.

\begin{figure}
\centering
\begin{subfigure}[b]{0.45\textwidth}
\includegraphics[width=\linewidth]{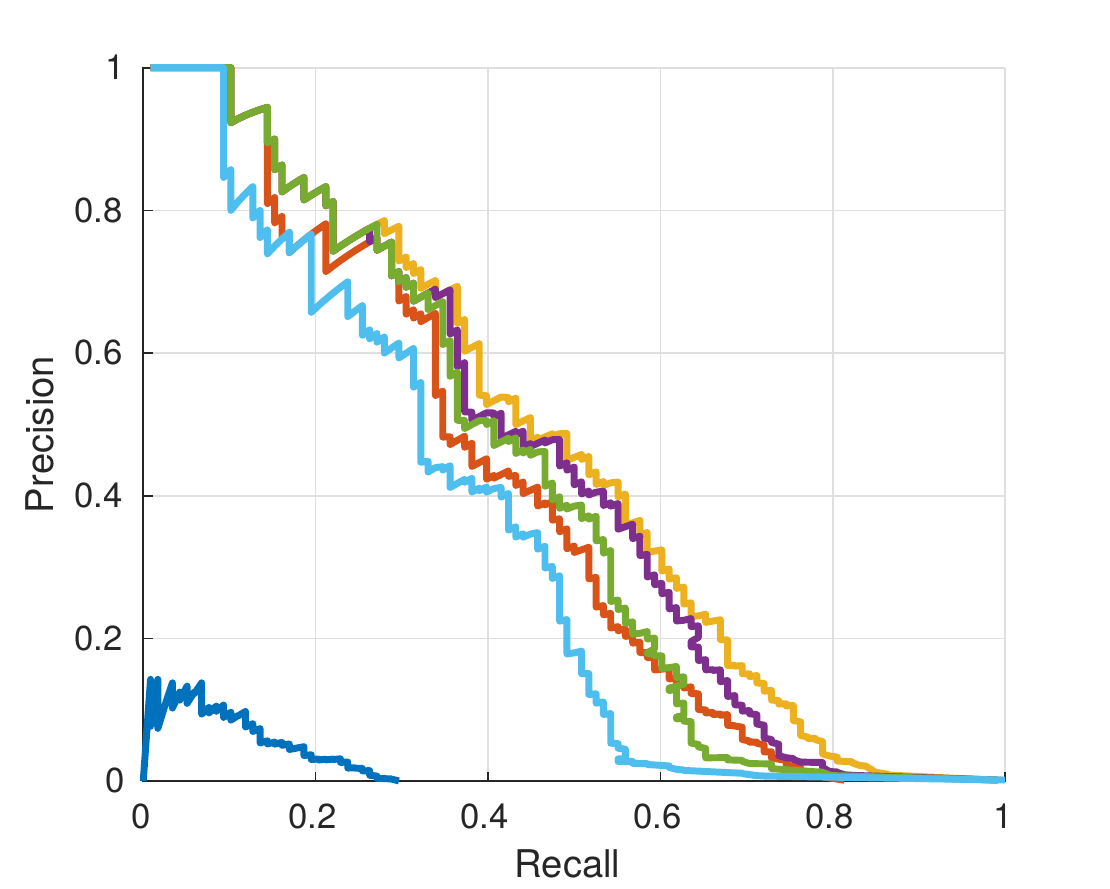}
\caption{Baseline}
\end{subfigure}
\begin{subfigure}[b]{0.45\textwidth}
\includegraphics[width=\linewidth]{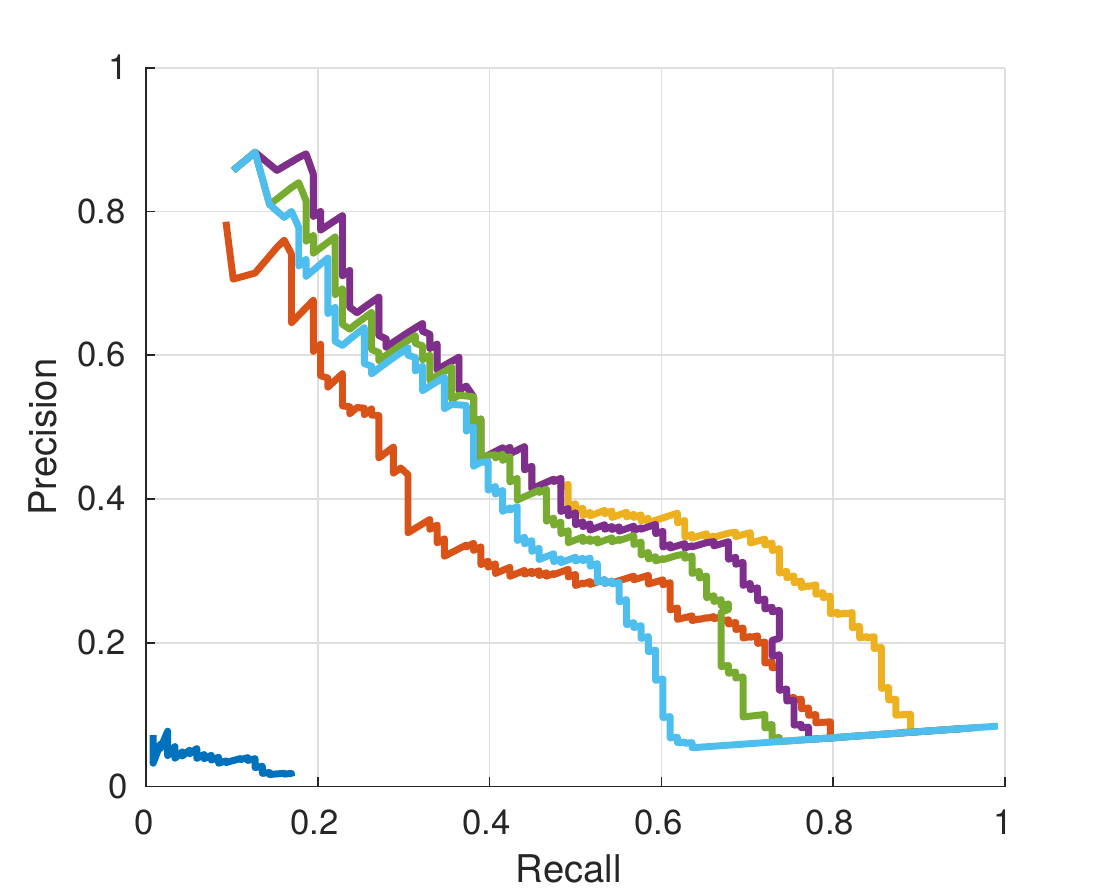}
\caption{CNN}
\end{subfigure}
\begin{subfigure}[b]{\textwidth}
\centering
\fbox{\includegraphics[width=0.4\linewidth]{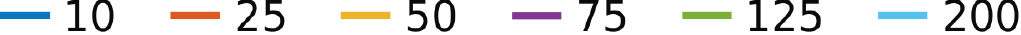}}
\caption{Legend (distance thresholds in pixels)}
\end{subfigure}
\caption{Precision-recall curves for the baseline (left) and CNN (right) on the validation set, evaluated with different distance thresholds. Too small thresholds like 10 pixels (40cm; dark blue) require a high positional precision which neither model can deliver; too large thresholds like 200 pixels (8m; light blue) include too many false alarms.}
\label{fig:EvaluationDistanceThresholds}
\end{figure}

Figure \ref{fig:EvaluationDistanceThresholds} shows the precision-recall curves for both models on the validation set, evaluated at thresholds ranging from 10 pixels (0.4m) to 200 pixels (8m). Several common trends can be observed, such as the weak performance at the smallest threshold. This implies that both models struggle in positioning their detections precisely on the exact location of the ground truth point. While this can partially be attributed to imperfect model performance, another potential cause is the limited precision in the ground truth itself, which may come from imperfect animal center delineation, detections on shadows, and the like.

At larger thresholds, the protocol is more tolerant towards spatial shifts. Consequently, the curves improve significantly. The baseline seems to score best at around 50 pixels, whereas the CNN has the best precision at 25 in the interesting region of high recall values. Furthermore, the absolute precision values at identical recalls are higher in case of the CNN compared to the baseline. Both these circumstances indicate that the CNN manages to provide predictions that are spatially closer to the actual ground truth.

Once the evaluation threshold exceeds a certain limit, the protocol includes predictions that are simply too far away from an animal to be accounted as a valid prediction. At the same time, multiple predictions of the same animal are discarded as false positives, which has negative effects on the precision. A look at the figures confirms that the precision-recall curves start to drop as soon as the distance threshold gets unrealistically high.

Following this ablation study, we decided to use a distance of 50 pixels (around 2m) in the results presented in the next section.

\subsection{Results on the full dataset}
\label{sec:resultsFullDataset}

\subsubsection{Animal instance-based evaluation} \label{ssec:instanceEval}

We evaluated the full models on the test set and assessed their performances using the census-oriented evaluation protocol discussed in the last section. All hyperparameters were chosen according to the best performance on the validation set described above\tc{}{ (see Section~\ref{sec:modelsUsed} and Table~\ref{tab:CNNparameters} for details)}.

\begin{figure}[!t]
\centering
\includegraphics[width=12cm]{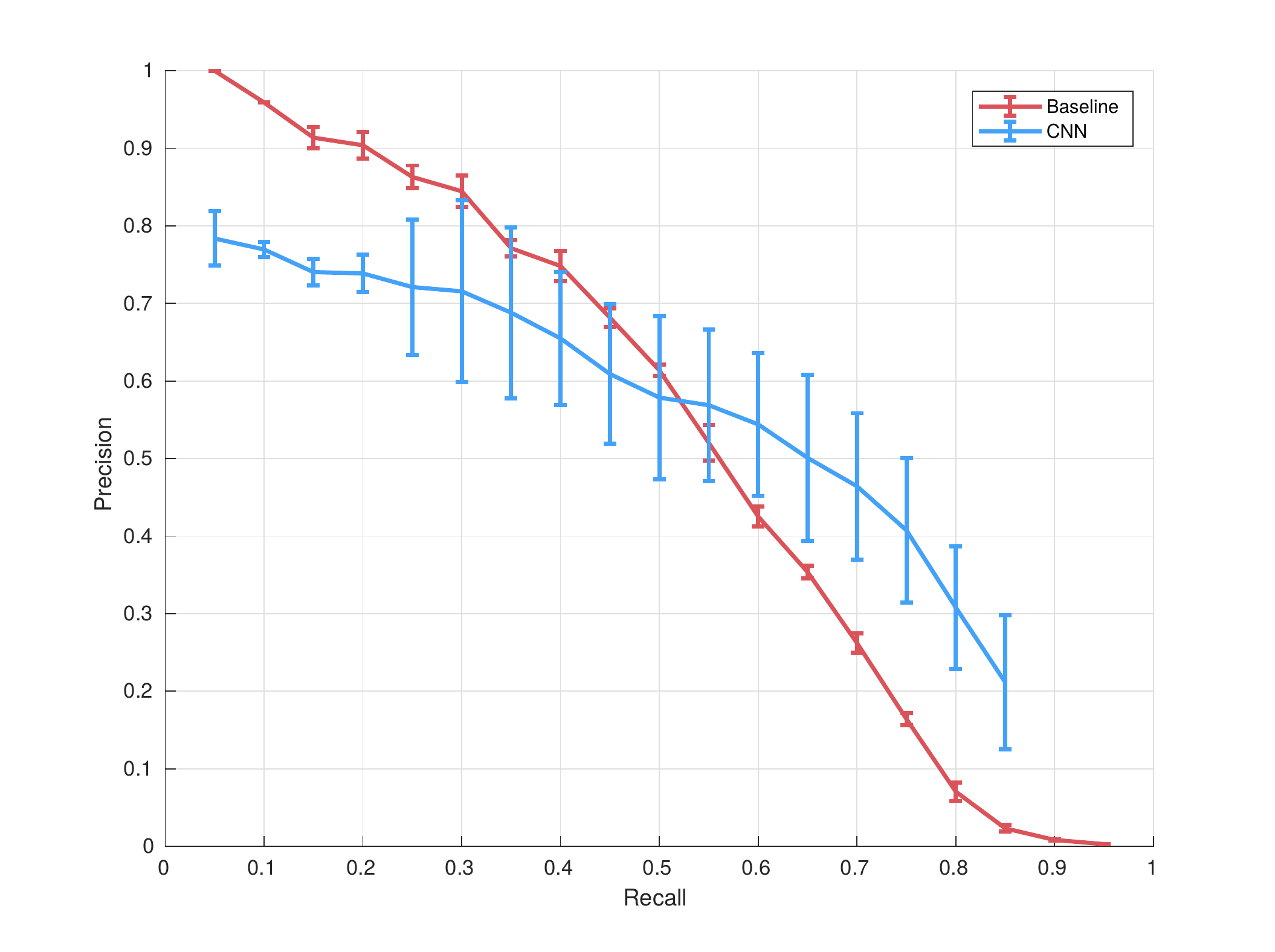}
\caption{Precision-Recall curves on the test set for the baseline (blue) and proposed CNN (purple), evaluated at a distance threshold of 50 pixels. \tc{}{Vertical bars denote the standard deviation for both models among the three cross-validation splits.} The proposed model manages to significantly reduce the number of false alarms at high recall rates.}
\label{fig:PRcurve_final_testSet}
\end{figure}

\begin{figure}
\centering
\includegraphics[width=\textwidth]{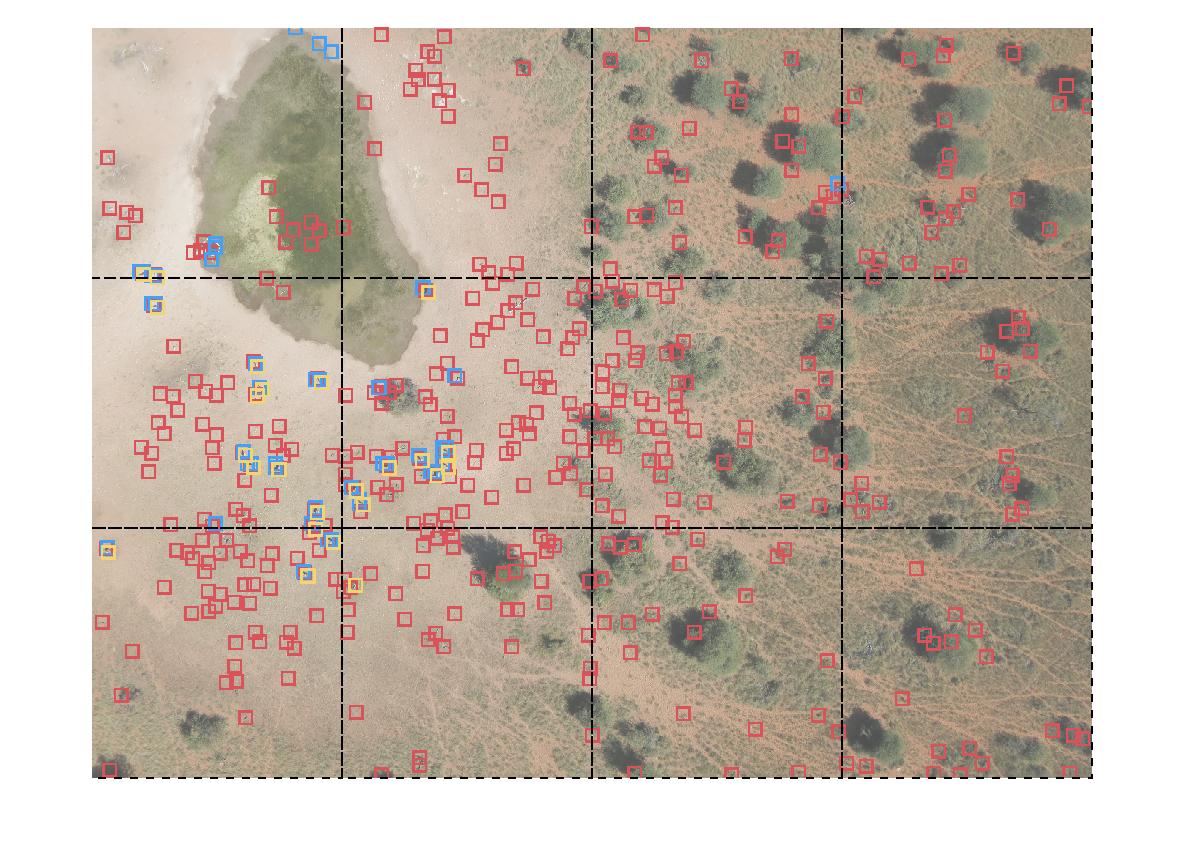}
\caption{Prediction results on a test set image. Both models were set to yield 90\% recall on the test set. The CNN (blue) manages to produce far less false alarms compared to the baseline (red). Ground truth locations are shown in yellow.}
\label{fig:examplePredictions_img}
\end{figure}

Figure \ref{fig:PRcurve_final_testSet} shows the obtained precision-recall curves for both models; Table \ref{tab:results_numeric} lists statistics for recall rates of 70 and 80\%, respectively. \tc{}{Values are provided for ranges reachable by all three respective cross-validated models. We note that the model variations along the three cross-validation splits are very low for the proposed CNN, and particularly controlled for the baseline. This indicates that any biases emerging from the choice of images in the dataset splits are virtually negligible.} Compared to the baseline, the CNN trained with the proposed recommendations (``Full model'') yields a significantly better precision at high recall values, which can be traced back to a greatly reduced number of false positives. In other words, the CNN yields similarly high recall values while making less mistakes. Numerically, the reduction in number of false positives is striking: from over \tc{20'000}{2500} to less than \tc{1'000}{450} if the target is to find \tc{90\%}{80\%} of the animals (Table \ref{tab:results_numeric}). \tc{}{For a recall level of 90\%, our full model produces 870 detections, whereas the baseline scores a staggering 20'688.} At such high recall values, it is unavoidable that models will struggle and provide false alarms, especially given the problems induced by covering large areas. However, looking at the results in a visual example (Figure \ref{fig:examplePredictions_img}), it becomes clear that the proposed CNN detects most animals limiting the mis-detections, while the baseline predicts animals almost everywhere, making the work of human annotators more time-consuming. On the contrary, the CNN detections can be rectified by a trained annotator in a limited amount of time.  At a target of 90\% recall, the baseline produces at least 51 and up to 773 detections per test image, even though 83 of the images do not contain animals at all. The CNN in turn predicts no animal correctly in 34 images and a maximum of only 81 detections in one image. In other words, this means that the CNN greatly reduces the effort required to verify the predictions.

\begin{table}
\centering
\begin{tabular}{c | c | c c  c | c c | c}
\hline
Recall & Model & TP & FP & FN & Precision & F1 & Num. tiles\\
\hline\hline
\multirow{2}{*}{0.7} & Baseline & 164.3 & 459.7 & 70.7 & 0.3 & 0.4 & 281.0\\
& CNN & 164.0 & 196.7 & 71.0 & 0.5 & 0.6 & 187.0\\
\hline
\multirow{2}{*}{0.8} & Baseline & 188.3 & 2546.0 & 46.7 & 0.1 & 0.1 & 942.3\\
& CNN & 188.0 & 447.3 & 47.0 & 0.3 & 0.4 & 268.0\\
\hline
\end{tabular}
\caption{Model performances at different levels of recall\tc{}{, averaged over the three cross-validation splits} (TP = true positives; FP = false positives; FN = false negatives\tc{}{; F1 = harmonic mean of precision and recall}). The proposed CNN manages to substantially reduce the number of false alarms, and with that the number of tiles to be screened (`Num. tiles'). Note that due to the confidence behavior of both models \tc{}{and the different splits} the exact numbers of true positives slightly differ.}
\label{tab:results_numeric}
\end{table}

\subsubsection{Tile-based evaluation}

In manual verification settings, the raw UAV image size ($\mathrm{4000\times3000}$ pixels) prohibits a complete animal identification by naked eye, given the small target size of only a few pixels (Figure \ref{fig:needleInHaystack}). It is thus common practice to split up the image into regular tiles and assess each one individually. If we assume a tile size of $\mathrm{1000\times1000}$ pixels as a realistic size for naked eye screening, the human operator would need to screen 12 tiles per full image acquired. Based on this assumption, we can calculate further statistics tailored towards model aptness for human post-processing. A good model will only predict locations in tiles that actually do contain an animal; less precise models will also fire in tiles without animals. This also means that we have hypothetical boundaries for a perfect model (a model that only scores in the tiles containing animals) and a lower bound (a model that scores in all tiles). In our test set, these figures lie at 97 and 1536 tiles, respectively.

\begin{figure}[!t]
\centering
\includegraphics[width=12cm]{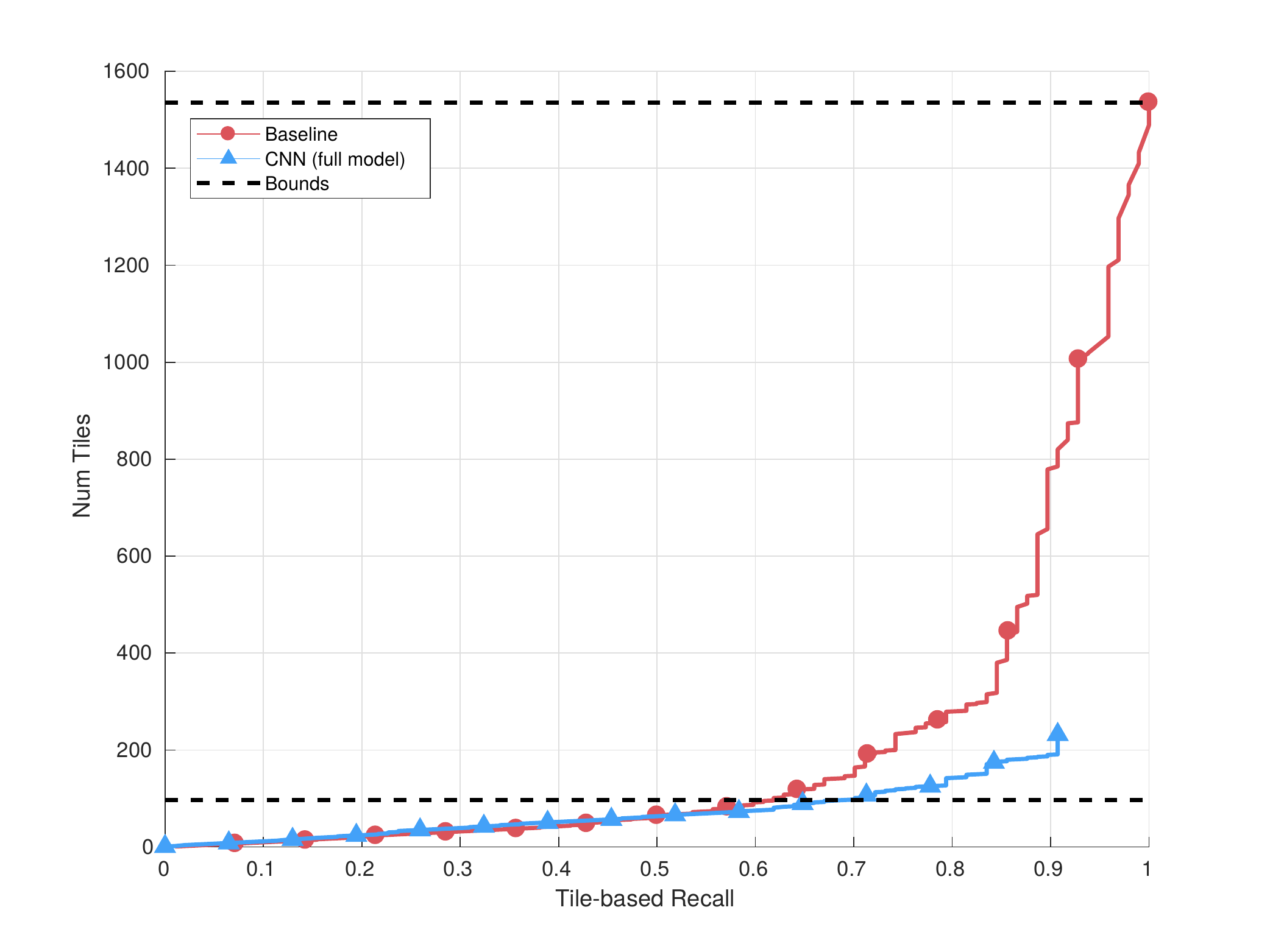}
\caption{Number of $\mathrm{1000\times1000}$ tiles with detections in comparison to a tile-based recall, given for both models on the test set. The total number of tiles in the test set was 1536 (upper dashed line); ground truth objects were found in 97 tiles (lower dashed line).}
\label{fig:p_r_tileBased}
\end{figure}

Figure \ref{fig:p_r_tileBased} shows the number of tiles versus tile-based recall for both the baseline and the CNN. Specific results at fixed recall values are provided in Table \ref{tab:results_numeric_tileBased}. On average, the proposed CNN detects animals in a lower number of tiles, while being able to concentrate predictions to the relevant ones (high tile-based recall). For instance, the CNN manages to find 90\% of the tiles containing animals (tile-based recall of 0.9), while only producing a total number of 190 tiles that need to be verified (93 more than theoretically needed). The baseline detects spurious animals in 779 out of the hypothetical 1536 tiles instead.

\begin{table}[!b]
\centering
\begin{tabular}{c | c | c c  c | c c | c}
\hline
Recall & Model & TP & FP & FN & Precision & F1 & Num. tiles\\
\hline\hline
\multirow{2}{*}{0.7} & Baseline & 68.0 & 95.0 & 29.0 & 0.4 & 0.5 & 163.0\\
& CNN & 68.0 & 41.0 & 29.0 & 0.6 & 0.7 & 109.0\\
\hline
\multirow{2}{*}{0.8} & Baseline & 77.7 & 200.0 & 19.3 & 0.3 & 0.4 & 277.7\\
& CNN & 78.0 & 77.0 & 19.0 & 0.5 & 0.6 & 155.0\\
\hline
\end{tabular}
\caption{Tile-based results at different recall levels\tc{}{, averaged over the cross-validation splits}. Tiles may encompass more than one ground truth animal; this results in overall higher precision scores compared to the individual animals evaluation presented in Section~\ref{ssec:instanceEval}.}
\label{tab:results_numeric_tileBased}
\end{table}

\begin{figure}[!t]
\centering
\includegraphics[width=12cm]{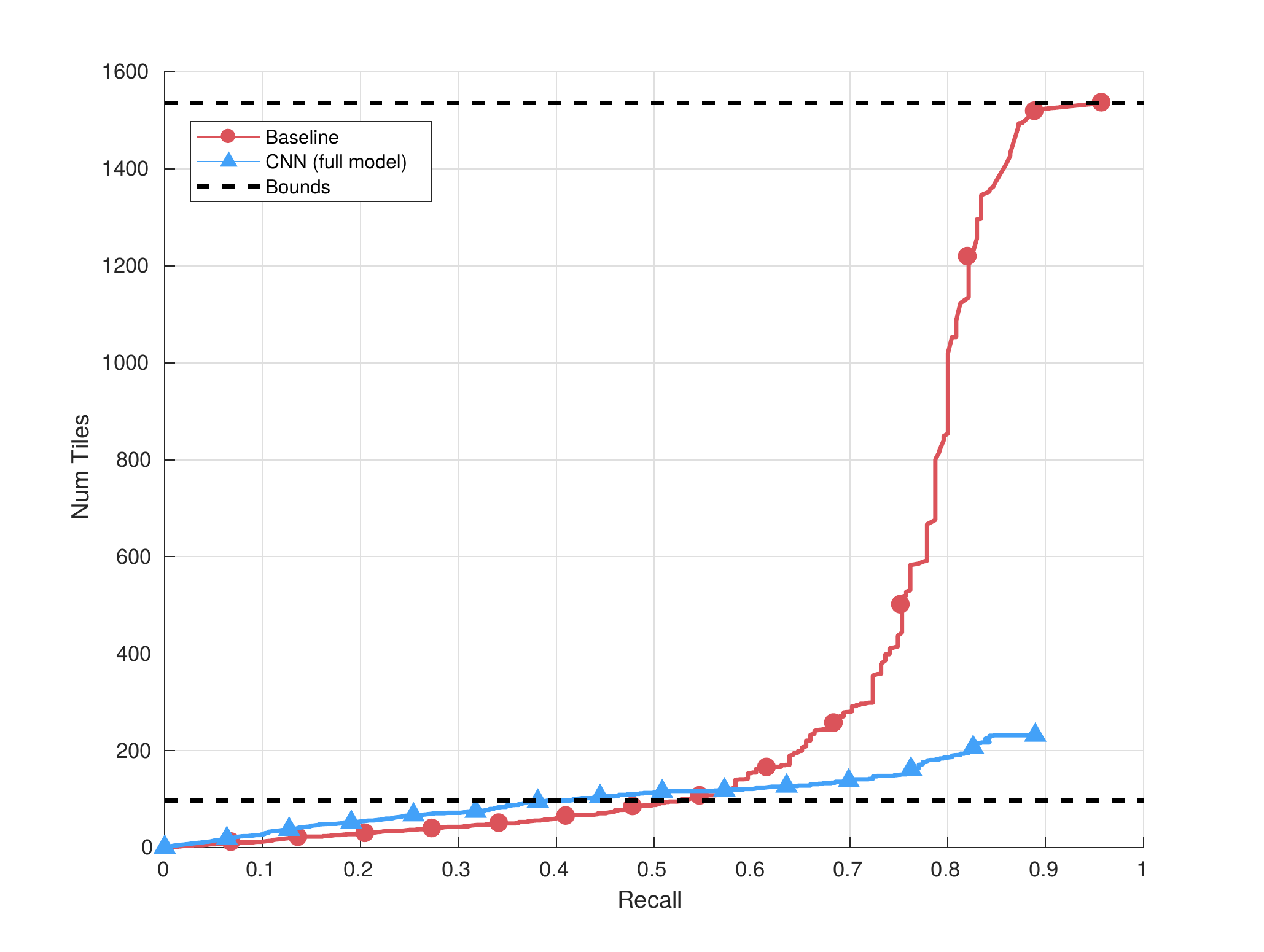}
\caption{Number of $1000\times1000$ tiles with detection for a given recall, reported on the test set.}
\label{fig:numTiles_vs_animalRecall}
\end{figure}

\begin{figure}[!t]
\centering
\includegraphics[width=\textwidth]{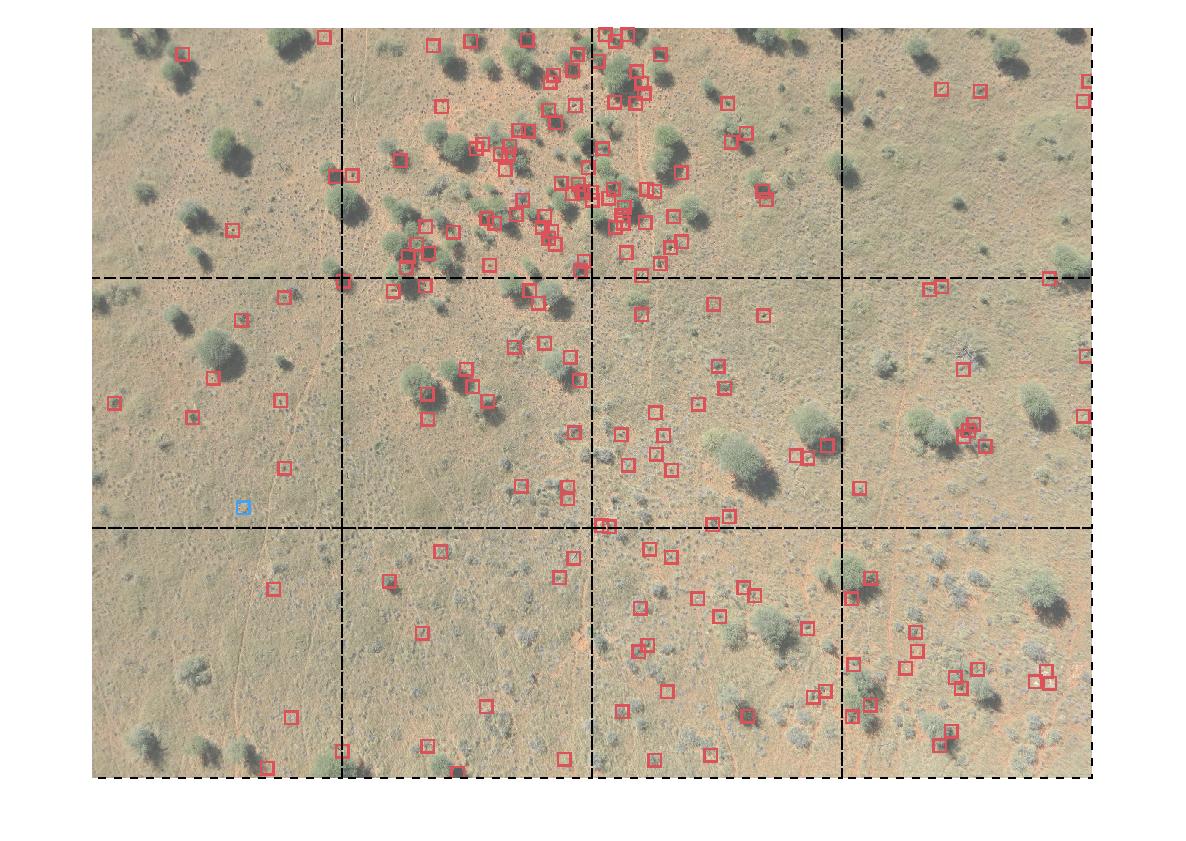}
\caption{Predictions on a test image that does not contain animals. Both the baseline (red) and CNN (blue) were set to yield a test set animal recall of 90\%. As can be seen, both models produce false alarms, but the baseline predictions are scattered across seven $\mathrm{1000\times1000}$ tiles (dashed) compared to the single tile covered by the CNN.}
\label{fig:examplePredictions_tileBased_img}
\end{figure}

Finally, Figure \ref{fig:numTiles_vs_animalRecall} shows the number of tiles against recall, based on the individual animals. This case can be seen as a combination on the annotator's target (i.e., finding all animals) and the way to get there (verifying a certain set of tiles). Although the curves share similarities with those in Figure \ref{fig:p_r_tileBased}, there are still notable differences. In particular, the CNN curve \tc{}{slightly} plateaus in the number of tiles at \tc{}{high} recall levels, which indicates that as soon as a certain threshold is exceeded, additional detections do not spread out more in space. At such recall values, the baseline produces at least one prediction in almost every tile, which makes the curve reach the hypothetical maximum of 1536 tiles. Visualizing one of the test images (Figure \ref{fig:examplePredictions_tileBased_img}) confirms this intuition. Summing up, the proposed CNN limits its detection to the relevant image tiles, thus easing the workload of human operators.

\section{Conclusion}

In this paper we discussed semi-automated models that work on UAV images and assist human operators in counting large mammals over vast areas, such as the African Savanna. This task is of central importance to animal conservation, as manual counting and photo-interpretation are prohibitively time-consuming. We focused in particular on the problem caused by the scarcity of animals with respect to the amount of images acquired in a campaign. Even a good model, one detecting with high precision on a selected subset of the data that is known to contain animals, tends to provide an enormous amount of false positives when applied to parts of the dataset containing swaths of empty savanna.

We proposed and discussed a methodology for animal censuses based on a Convolutional Neural Network (CNN) and showed how to train deep animal detectors on a real-world UAV image dataset consisting mostly of images with no animals. We introduced several recommendations to guide a CNN in the right direction during training. These include \emph{(i.)} class-weighting to soften the impact of the overly abundant background class, \emph{(ii.)} curriculum learning to prevent the model from forgetting animal representations, \emph{(iii.)} hard negative mining to further boost the precision, and \emph{(iv.)} the introduction of a border class around animals that alleviates the confusion of nearby animals over background locations. Through an ablation study we demonstrated that every technique is necessary to reach the best performances.

In census applications, a pixel-precise localization of animals is only of secondary interest. To account for this fact, we proposed an alternative evaluation protocol that assesses model performances based on the number of detections they provide (compared to the actual number of targets), but poses less restrictions on the positional accuracy of predictions. For cases where animal detectors are used as a pre-selection step for a following manual verification stage, we further presented a second evaluation method that accounts for the number of image tiles a detector finds animals in. A good detector will only detect in the set of tiles that contain actual animals; a bottom baseline will require all tiles in the full dataset to be screened.

We demonstrated the effectiveness of all our recommendations in an experiment in the Namibian game reserve of Kuzikus, where we trained a CNN according to the presented recommendations and compared it to the state-of-the-art using the two evaluation protocols. The results show that the CNN not only manages to yield a substantially higher precision at high recall values when compared to a state-of-the-art detector, but also manages to produce more confined predictions, spreading across a lower number of image tiles. In practice, this results in the CNN lowering the number of image tiles to be verified to less than one third compared to the baseline, at a recall level of 90\%. Our training recommendations are model-agnostic and straight-forward to apply to any deep learning-based object detector, and the two evaluation protocols complement the model assessment. In sum, both parts demonstrate the necessity and a potential path to go towards animal censuses that are actually achievable during a realistic UAV campaign containing countless images with no wildlife recorded.


\section*{Acknowledgements}

This work has been supported by the Swiss National Science Foundation (grant PZ00P2-136827 (DT, \url{http://p3.snf.ch/project-136827}). The authors would like to acknowledge the SAVMAP consortium (in particular Dr. Friedrich Reinhard of Kuzikus Wildlife Reserve, Namibia) and the QCRI and Micromappers (in particular Dr. Ferda Ofli and Ji Kim Lucas) for the support in the collection of ground truth data.

\section{References}

\bibliographystyle{elsarticle-harv}
\biboptions{authoryear}

\bibliography{ms}


\end{document}